\documentclass[lettersize,journal]{IEEEtran}
\usepackage{amsmath,amsfonts}
\usepackage{array}
\usepackage[caption=false,font=footnotesize]{subfig}
\usepackage{textcomp}
\usepackage{stfloats}
\usepackage{url}
\usepackage{verbatim}
\usepackage{graphicx}
\usepackage{multirow}
\usepackage[table,xcdraw]{xcolor}
\usepackage[numbers,sort&compress]{natbib}
\usepackage{natbib}
\usepackage{algorithm}

\usepackage{xspace}
\usepackage{makecell}
\usepackage{colortbl}
\usepackage{amsmath}
\usepackage{amsthm}
\usepackage{booktabs}
\usepackage{pifont}
\usepackage{bbding}
\usepackage{subcaption}
\usepackage{amssymb}

\usepackage{bbding} 
\usepackage[noend]{algpseudocode}
\usepackage{siunitx}
\usepackage{pgfplots}
\pgfplotsset{compat=1.17}
\usepackage{footnote} 

\newcommand{\methodname}{UAPA}

\usepackage{algpseudocode} 
\usepackage{bm}
\usepackage{booktabs} 
\usepackage{arydshln}
\usepackage{wrapfig}

\usepackage{tcolorbox}
\tcbuselibrary{skins, breakable, listings}
\newtcolorbox[auto counter]{mybox}[1][]{
  title=My Box \thetcbcounter,
  label=mybox:\thetcbcounter,
  #1
}
\usepackage{tabularx}
\usepackage{etoc}
\usepackage{hyperref}
\definecolor{myblue}{RGB}{47,85,151}
\definecolor{myred}{RGB}{192,80,77}
\definecolor{myshade}{RGB}{238,240,244}
\definecolor{mygrid}{RGB}{220,220,220}

\begin{document}

\title{Uncertainty-Aware Prototype Learning for Open-Set Test-Time Adaptation}

\author{Jinwu Hu$^*$, Langyu Bian$^*$, Jian Sun, Dongjin Yang, Chao Shuai, Yufeng Wang, Kai Zhou, Bin Xiao$^\dag$, \\ Mingkui Tan$^\dag$,~\IEEEmembership{Member,~IEEE}, Hongmin Cai$^\dag$, \IEEEmembership{Senior Member, IEEE}, Wei Luo$^\dag$

\thanks{Jinwu Hu, Langyu Bian, Jian Sun, Dongjin Yang, and Kai Zhou are with the School of Software Engineering, South China University of Technology, Guangzhou, China (E-mail: fhujinwu@gmail.com, blangy2003@gmail.com, jiansun97@163.com, dongjinyang2003@gmail.com, kayjoe0723@gmail.com).}
\thanks{Chao Shuai is with Zhejiang University, Hangzhou, China (E-mail: chaoshuai@zju.edu.cn).}
\thanks{Yufeng Wang and Hongmin Cai are with the School of Future Technology, South China University of Technology, Guangzhou, China (E-mail: yufeng6568@gmail.com,  hmcai@scut.edu.cn).}
\thanks{Bin Xiao is with the Chongqing University of Posts and Telecommunications, Chongqing, China (E-mail: xiaobin@cqupt.edu.cn).}
\thanks{Mingkui Tan is with the School of Software Engineering, South China University of Technology, and also with the Key Laboratory of Big Data and Intelligent Robot, Ministry of Education, Guangzhou, China (E-mail: mingkuitan@scut.edu.cn).}
\thanks{Wei Luo is with the South China Agricultural University (E-mail: cswluo@gmail.com).}
\thanks{$^*$Authors contributed equally. $^\dag$Corresponding authors}
}

\maketitle

\begin{abstract}
Open-set test-time adaptation (TTA) aims to adapt source-trained models to unlabeled test streams where covariate-shifted in-distribution (csID) samples and covariate-shifted out-of-distribution (csOOD) samples may coexist. Existing TTA methods are typically designed under a closed-set assumption and therefore lack explicit mechanisms for reliably distinguishing csID from csOOD samples. Consequently, csOOD samples may be mistakenly included in confidence-enhancing adaptation, producing harmful gradients that degrade known-class performance. To address this, we propose \textbf{U}ncertainty-\textbf{A}ware \textbf{P}rototype Learning for Open-Set Test-Time \textbf{A}daptation (\textbf{\methodname}), a \textbf{plug-and-play} method for reliable open-set adaptation. Specifically, we first introduce Uncertainty-Driven ID/OOD Separation, which combines a classifier-prototype score with a mini-batch-normalized evidence-derived uncertainty score to distinguish csID from csOOD samples. We then propose a Dynamic Prototype Refinement strategy that periodically refines class-specific target prototypes using buffered csID features to progressively follow the estimated class directions. Lastly, we introduce Contrastive Prototype Entropy Adaptation, which applies entropy minimization to csID samples and entropy maximization to csOOD samples, while prototype-level contrastive learning aligns csID features with their corresponding target prototypes and reduces the maximum similarity between csOOD features and known-class prototypes. Empirically, UAPA outperforms the strongest baselines on the CIFAR benchmarks and Tiny-ImageNet-C, improving average OSCR by 2.08 and 1.28 percentage points, respectively. 
\end{abstract}


\section{Introduction}

\IEEEPARstart{D}{eep} neural networks have achieved remarkable progress in visual recognition tasks~\cite{he2016deep, dosovitskiyimage,ContiFMRWR26,WuLSYNL25,zhuangweitpami2026,WangHGWL26}, yet their reliability can substantially degrade when deployed in open-world environments~\cite{ZhouLQXL23,FengXT21,WangXTGTSB26,DuWSWLZDWWL24}. In real-world test streams, known-class samples may undergo appearance-level changes caused by weather variations, motion blur, sensor noise, or other visual corruptions while retaining the source label space, a setting known as \emph{covariate shift}~\cite{koh2021wilds}. Meanwhile, test samples may also come from classes outside the source label space, leading to \emph{semantic shift}~\cite{geng2020recent}. In practice, these two shifts are often coupled: unknown-class samples may appear under the same degraded conditions as shifted known-class samples. For example, an autonomous driving system may encounter familiar vehicles under fog or motion blur, while also observing unseen obstacles or rare objects that were absent during training. A reliable deployed model is therefore expected not only to adapt to shifted known classes, but also to avoid being misled by shifted unknown classes~\cite{ShiH0022,ZhengGX16,gao2024unified,zhao2026mitigating,zhang2026learning}.

\begin{figure}[t]
\centering
\includegraphics[width=1.0\linewidth]{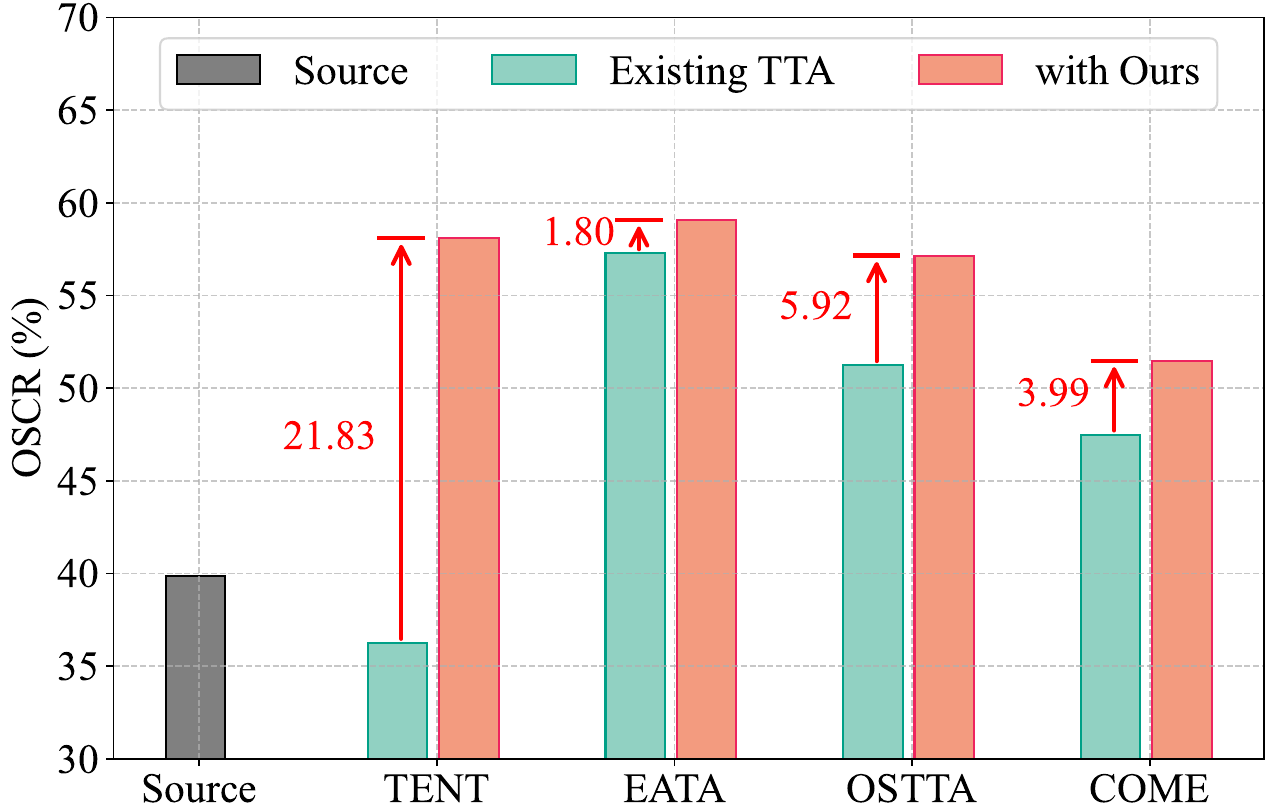}
\caption{OSCR comparison on CIFAR-100-C before and after incorporating UAPA into representative TTA methods~\cite{wang2021tent,niu2022eata,lee2023towards,zhangcome}. Values above the arrows denote absolute OSCR gains in percentage points.}
\label{fig:existing_with-ours}
\end{figure}

To improve robustness under covariate shift, test-time adaptation (TTA) updates a source-trained model online using unlabeled test data~\cite{wang2021tent,niu2022eata,niutowards,zhangcome,tan2025uncertainty}. Representative TTA methods combine unsupervised adaptation objectives, such as entropy minimization~\cite{wang2021tent,niu2022eata} or prediction consistency~\cite{tan2025uncertainty,chen2026zerosiam}, with update strategies such as reliability-based sample selection~\cite{niu2022eata,niutowards}. These designs enable the model to adapt to shifted test distributions without requiring access to the original source data. However, many TTA methods are developed under an implicit closed-set assumption, where each test sample is assumed to belong to one of the source classes despite possible changes in the test distribution. Under this assumption, encouraging confident predictions or selecting reliable samples can help the model adapt to shifted known-class samples.

In open-set test streams, this closed-set assumption no longer holds because the target stream may contain both known-class samples with appearance or quality shifts, referred to as covariate-shifted in-distribution (csID) samples, and unknown-class samples outside the source label space under similar shifts, referred to as covariate-shifted out-of-distribution (csOOD) samples~\cite{gao2024unified,zhao2026mitigating,XiaHWDLML23}. This setting raises two key challenges. \textit{1) Negative adaptation caused by csOOD samples.} Entropy minimization encourages confident predictions for target samples. If csOOD samples are adapted in the same way as csID samples, they may be pushed toward overconfident known-class predictions, yielding harmful gradients that distort decision boundaries and cause negative adaptation. \textit{2) Unreliable separation between csID and csOOD samples.} The model must decide which samples can support known-class adaptation and which should be excluded from confidence-enhancing updates. This is difficult because severe covariate shifts can make csID samples uncertain, whereas some csOOD samples may appear confident. Thus, relying on a single confidence, entropy, or prototype-similarity score~\cite{gao2024unified} can produce unreliable partitions and unstable adaptation.

To address these challenges, we propose \textbf{U}ncertainty-\textbf{A}ware \textbf{P}rototype Learning for Open-Set Test-Time \textbf{A}daptation (\textbf{\methodname}), a \textbf{plug-and-play} method (see Fig. \ref{fig:existing_with-ours}) for reliability-aware adaptation in open-set test streams. The key idea of \methodname{} is to determine the adaptation role of each test sample according to both its geometric relation to known classes and an evidence-derived uncertainty score computed from the online-model logits. Specifically, \methodname{} first introduces Uncertainty-Driven ID/OOD Separation, which combines a classifier-prototype score with a mini-batch-normalized evidence-derived uncertainty score to partition incoming samples into csID and csOOD subsets. It then performs Dynamic Prototype Refinement (DPR), which accumulates selected csID features in class-wise buffers and periodically updates class-specific target prototypes. 
Finally, \methodname{}  develops Contrastive Prototype Entropy Adaptation (CPEA), which applies different entropy objectives to the estimated csID and csOOD subsets and uses progressively initialized target prototypes to provide stage-dependent, prototype-relative regularization. 
In this way, \methodname{} jointly performs sample separation, prototype maintenance, and direction-aware adaptation to suppress negative adaptation in open-set test-time adaptation.
Our main contributions are as follows:
\begin{itemize}
    \item We propose an uncertainty-driven ID/OOD separation mechanism for open-set test-time adaptation. By jointly considering a classifier-prototype score and a mini-batch-normalized evidence-derived uncertainty score, \methodname{} incorporates complementary geometric and uncertainty-related cues for csID/csOOD partitioning. We characterize the conditional separability and bounded-perturbation stability of the fused score, while clarifying the mini-batch dependence of the normalized uncertainty score and its distinction from GMM-based posterior gating.
    \item We propose Dynamic Prototype Refinement and Contrastive Prototype Entropy Adaptation for open-set test-time adaptation. DPR periodically updates class-specific target prototypes using buffered csID features, while CPEA combines csID entropy minimization, csOOD entropy maximization, and prototype-level contrastive learning to suppress negative adaptation. We show that DPR induces one-step angular non-expansion toward the buffered class direction, while CPEA encourages pseudo-label-relative margin enlargement for selected csID samples and nearest-prototype separation for selected csOOD samples in the hyperspherical feature space.
    \item Extensive experiments on CIFAR-C and Tiny-ImageNet-C benchmarks demonstrate that~\methodname{} consistently improves open-set classification performance across different backbones and state-of-the-art TTA baselines. Further plug-and-play and efficiency analyses show that~\methodname{} can be integrated into existing TTA methods with low additional computational cost.
\end{itemize}

\section{Related Work}
\label{sec:related work}

\subsection{Test-Time Adaptation}
Test-time adaptation (TTA)~\cite{wang2021tent, niu2022eata, niutowards, zhangcome, niu2025selfbootstrapping,yang2024test,fuchs2025online,deng2025learning,zhou2025bayesian,deng2025test,hutest,zhang2026learning} adapts a source-trained model during inference without accessing source data, using lightweight online updates that often restrict learning to normalization statistics or a small subset of parameters. It differs from test-time training (TTT), which performs an auxiliary self-supervised update before prediction and requires training-time modifications~\cite{sun2020test}. A foundational approach is entropy minimization, as in TENT~\cite{wang2021tent}, which updates the model batch by batch to increase prediction confidence under covariate or domain shift. EATA~\cite{niu2022eata} selects reliable, non-redundant samples to avoid harmful updates and forgetting. COME~\cite{zhangcome} replaces plain entropy minimization with a conservative variant grounded in evidential or Dirichlet uncertainty to curb overconfidence. Self-Bootstrapping~\cite{niu2025selfbootstrapping} proposes a versatile consistency objective that extends TTA beyond classification to tasks such as segmentation and detection. A$^2$Memory~\cite{zhang2026learning} improves memory-based online TTA for vision-language models via attribute-centric soft assignment and class-wise associative memory, reducing memory contamination and enabling sample-adaptive prototype retrieval. Despite these advances, most TTA methods are designed for closed-set covariate or domain shifts, where test samples are assumed to belong to the source label space.

\subsection{Open-Set Test-Time Adaptation}
Open-set test-time adaptation (OSTTA)~\cite{lee2023towards,gao2024unified,yu2024stamp,dong2025towards,wang2025activate,sreenivasefficient} considers target streams that include unknown classes in addition to covariate-shifted known classes. Closed-set TTA often fails in this setting because uniform entropy minimization treats unknown inputs as trustworthy, which pushes them toward over-confident known predictions and degrades the decision boundary~\cite{lee2023towards}. A common strategy is to separate target samples into covariate-shifted in-distribution (csID) and covariate-shifted out-of-distribution (csOOD) and to assign different updates. UniEnt~\cite{gao2024unified} performs mixture-based posterior gating on a prototype-similarity or uncertainty score to mine pseudo-csID and pseudo-csOOD, then minimizes entropy for the former and maximizes entropy for the latter, with a weighted variant to reduce partition noise. STAMP~\cite{yu2024stamp} avoids risky mini-batch updates by optimizing over a stable memory bank that is refreshed with low-entropy and label-consistent samples, improving robustness under streaming conditions. AEO~\cite{dong2025towards} proposes adaptive entropy-aware optimization for multimodal OSTTA, enlarging the entropy gap between known and unknown samples to better identify unknown classes under multimodal distribution shifts. ROSETTA~\cite{zhao2026mitigating} studies the ID-OOD trade-off in OSTTA and aims to improve csOOD detection while preserving csID classification performance.

\subsection{Out-of-Distribution Detection}
Out-of-distribution (OOD) detection aims to identify samples that deviate from the training distribution, preventing overconfident predictions on unseen or abnormal inputs~\cite{hendrycks2017baseline,liang2018enhancing,liu2020energy,ChengJL25,WangHLGLL25,HuWLLH26}. Maximum Softmax Probability (MSP)~\cite{hendrycks2017baseline} uses the maximum softmax probability as a confidence score, providing a baseline for detecting misclassified and OOD samples. ODIN~\cite{liang2018enhancing} introduces temperature scaling and input perturbations to enlarge the confidence gap between in-distribution and OOD samples, improving the reliability of softmax-based detection. Energy-based OOD detection~\cite{liu2020energy} replaces softmax confidence with an energy score derived from logits, addressing the limitation that softmax probabilities may be poorly calibrated for OOD inputs. PRO~\cite{chen2025leveraging} leverages perturbation robustness to construct more discriminative OOD scores, based on the observation that OOD confidence is more sensitive to adversarial perturbations. 
W-DOE~\cite{WangHLGLL25} improves open-world OOD generalization by expanding auxiliary OOD coverage through implicit data synthesis and distribution-agnostic optimization.
Unlike conventional OOD detection methods that mainly focus on post-hoc scoring for a fixed model, \methodname{} integrates OOD estimation with online adaptation, using prototype-based similarity and a mini-batch-normalized evidence-derived uncertainty score to separate csID/csOOD samples and suppress csOOD-induced negative adaptation.

\subsection{Contrastive Learning}
Contrastive learning learns discriminative representations by pulling semantically similar samples closer and pushing dissimilar samples apart in the embedding space~\cite{DuWSH24,LiWZJDLXT25,ZhengYWQZWX24,chen2020simple,he2020momentum,grill2020bootstrap,chen2021exploring,WuYGDJN23}. SimCLR~\cite{chen2020simple} demonstrates that strong data augmentation, large batch sizes, and nonlinear projection heads enable contrastive representation learning. MoCo~\cite{he2020momentum} introduces a momentum encoder and a dynamic queue to maintain a large and consistent dictionary of negative samples, alleviating the dependence on extremely large batches. Beyond negative-sample-based methods, BYOL~\cite{grill2020bootstrap} and SimSiam~\cite{chen2021exploring} adopt asymmetric Siamese architectures with prediction heads and stop-gradient operations, enabling representation learning without explicit negative pairs. ReSSL~\cite{ZhengYWQZWX24} extends instance-level contrastive learning by modeling inter-instance relations through similarity-distribution matching under weak augmentations. CHMatch~\cite{WuYGDJN23} combines robust adaptive thresholding with contrastive hierarchical matching to select reliable samples and learn discriminative representations.

\begin{figure*}[t]
\centering
\includegraphics[width=0.97\linewidth]{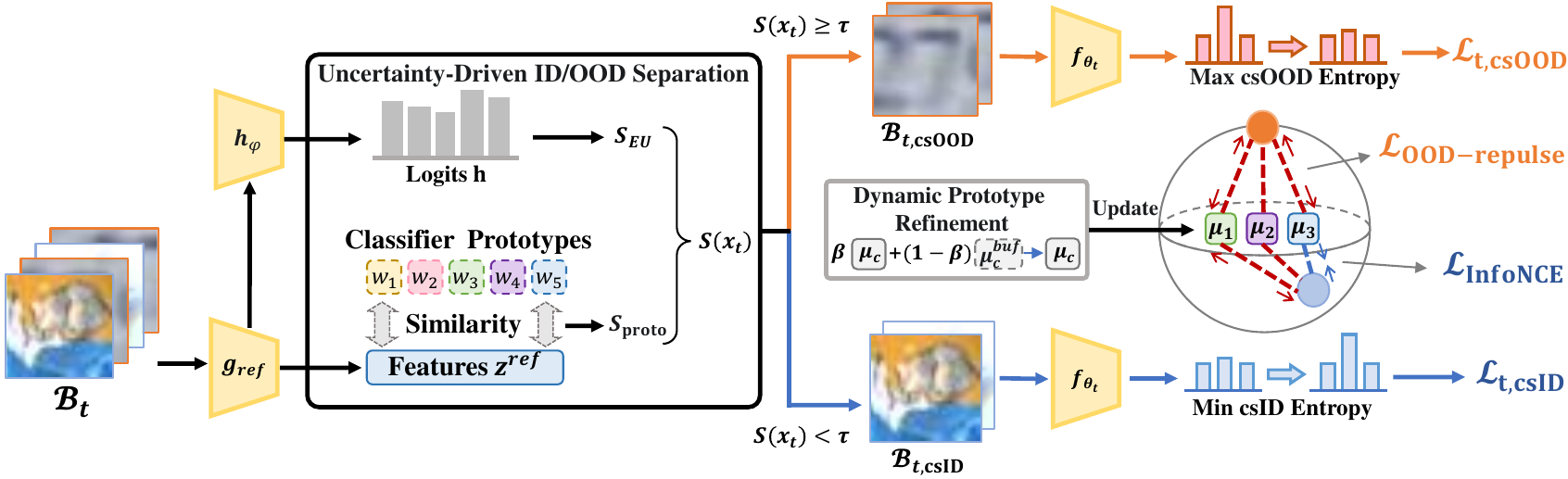}
\caption{Overview of \methodname{}. For each incoming batch $\mathcal{B}_t$, \methodname{} combines the classifier-prototype score $S_{\mathrm{proto}}$ with the mini-batch-normalized evidence-derived uncertainty score $S_{\mathrm{EU}}$ to compute a separation score $S(x_t)$, which partitions samples into csID and csOOD subsets using threshold $\tau$. Contrastive Prototype Entropy Adaptation applies entropy minimization to csID samples, entropy maximization to csOOD samples, and prototype-level contrastive learning to regularize feature representations. Dynamic Prototype Refinement periodically updates class-specific target prototypes using buffered csID features.}
\label{fig:method}
\end{figure*}

\section{Problem Formulation}
\label{sec:problem_statement}

Let $P(x)$ denote the distribution of the source training samples
$\{x_i^s\}_{i=1}^{N_s}$, where $x_i^s \sim P(x)$. The \textbf{base model} $f_{\theta^{\circ}}(x)$ is trained on the labeled source dataset $\mathcal{D}_s = \{(x_i^s, y_i^s)\}_{i=1}^{N_s} \subset \mathcal{X}\times\mathcal{Y}_s$, where $\mathcal{Y}_s=\{1,\dots,\mathcal{C}_S\}$ is the set of $\mathcal{C}_S$ known classes.
After training, the model $f_{\theta^{\circ}}(x)$ is expected to fit
the source data distribution. At inference time, the model is expected to perform well on in-distribution test data, namely, $x \sim P(x)$. However, in real-world deployments, due to possible distribution shifts between training and test data, we may encounter target samples drawn from a shifted distribution $Q(x)$, where $Q(x)\neq P(x)$.
The types of distribution shifts that can occur during inference
are as follows. \textit{1) Covariate Shift:} The samples share the same set of classes but differ in their distribution due to variations in style or domain. \textit{2) Semantic Shift:} The target domain contains novel classes that were not present during training on the source domain. In this case, the source label space is a strict subset of the
target label space, \textit{i.e.}, $\mathcal{Y}_s \subset \mathcal{Y}_t$, where $\mathcal{Y}_t$ denotes the target label space.
These shifts partition the target data stream into two categories:
\textbf{1) Covariate-shifted In-Distribution (csID).} Samples $x^t \in \mathcal{B}_t$ whose true labels $y^t$ belong to the known source classes, \textit{i.e.}, $y^t \in \mathcal{Y}_s$. \textbf{2) Covariate-shifted Out-of-Distribution (csOOD).} Samples $x^t \in \mathcal{B}_t$ whose true labels $y^t$ are from novel, unknown classes, \textit{i.e.},
$y^t \in \mathcal{Y}_t \setminus \mathcal{Y}_s$. For csOOD samples, predictions over the known source classes can be unreliable, leading to poor open-set performance.

Test-time adaptation (TTA)~\cite{wang2021tent,niu2022eata,chentowards,zhangcome} seeks to improve model performance in the target domain by adapting the model using only test data. Specifically, given a set of target test samples
$\{x_j\}_{j=1}^M$, where $x_j \sim Q(x)$ and $Q(x) \neq P(x)$, one needs to adapt $f_{\theta}(x)$ to improve prediction performance
on the target data. To achieve this, many methods update the model by minimizing an unsupervised objective defined on test samples:
\begin{equation}
    \label{tta_object}
    \min_{\overline{\theta}} \mathcal{L}(x;\theta),
    \quad x \sim Q(x),
\end{equation}
where $\overline{\theta}$ denotes the subset of parameters in $\theta$ that is updated during TTA. The learning objective $\mathcal{L}(\cdot)$ is often formulated as an entropy minimization problem.

However, in open-set scenarios where both csID and csOOD samples coexist, traditional TTA methods struggle to perform. If the same closed-set adaptation objective is applied to all target
samples, csOOD samples may be forced toward overconfident predictions
over known classes and generate harmful gradients during online updates. As a result, they can distort decision boundaries, contaminate
adaptation statistics or prototypes, and degrade the performance
on csID samples. The goal of open-set TTA is therefore to improve classification on latent csID samples, identify latent csOOD samples, and prevent the latter from inducing harmful adaptation updates.

\begin{algorithm}[t]
\caption{Pipeline of the proposed \methodname{}}
\label{alg:our_method}
\small
\begin{algorithmic}[1]
    \State \textbf{Input:} Test stream $\mathcal{D}_{\mathrm{test}}$, source model $M_0$, batch size $B$, and hyperparameters $\lambda,\tau,\gamma,\alpha_0,\alpha_1, \rho,\beta,\eta,\tau_0,$ and $\epsilon$.
    \State \textbf{Initialize:} Online model $M_1\leftarrow M_0$, reference model $M_{\mathrm{ref}}\leftarrow\operatorname{Copy}(M_0)$, target prototypes $\mu_c\leftarrow\mathbf{0}$, feature buffers $\mathcal{F}_c\leftarrow\emptyset$ for all $c$, and effective-batch counter $k\leftarrow0$.

    \For{each incoming mini-batch $\mathcal{B}_t\subset\mathcal{D}_{\mathrm{test}}$}
        \State Obtain predictions and online features for samples in $\mathcal{B}_t$.
        \State Compute $S_{\mathrm{proto}}$ using $M_{\mathrm{ref}}$ and classifier prototypes.
        \State Compute the raw evidence-derived uncertainty score $U_{\mathrm{EU}}$ and its mini-batch-normalized score $S_{\mathrm{EU}}$ using online predictions.
        \State Compute $S=\lambda S_{\mathrm{proto}}+(1-\lambda)S_{\mathrm{EU}}$ and split $\mathcal{B}_t$ into $\mathcal{B}_{t,\mathrm{csID}}$ and $\mathcal{B}_{t,\mathrm{csOOD}}$ using threshold $\tau$.
        \State Store csID features in class-wise buffers according to their pseudo-labels.
        \State Compute $\mathcal{L}_{t,\mathrm{entropy}}$ and
        $\mathcal{L}_{t,\mathrm{contrast}}$.
        \State Update the online model by minimizing
        $\mathcal{L}_{t,\mathrm{total}}
        =
        \mathcal{L}_{t,\mathrm{entropy}}
        +
        \gamma\mathcal{L}_{t,\mathrm{contrast}}$.

        \If{$\mathcal{B}_{t,\mathrm{csID}}\neq\emptyset$}
            \State $k\leftarrow k+1$.
            \If{$k\bmod\eta=0$}
                \State Update each target prototype with a non-empty buffer using momentum $\beta$.
                \State Clear all class-wise feature buffers.
            \EndIf
        \EndIf
    \EndFor
    \State \textbf{Output:} Pre-update predictions for all test samples.
\end{algorithmic}
\end{algorithm}

\section{Uncertainty-Aware Prototype Learning}
In this paper, we propose \textbf{U}ncertainty-\textbf{A}ware \textbf{P}rototype Learning for Open-Set Test-Time \textbf{A}daptation (\textbf{\methodname}), a \textbf{plug-and-play method} for open-set TTA, which adapts the model using samples selected as csID while
limiting harmful updates from samples selected as csOOD. The overall framework is illustrated in Fig.~\ref{fig:method}, and the pipeline is summarized in Alg.~\ref{alg:our_method}. We first introduce Uncertainty-Driven ID/OOD Separation (\textit{cf.} Sec.~\ref{sec: peparation}), which combines a classifier-prototype score with a mini-batch-normalized evidence-derived uncertainty score to partition incoming samples into csID and csOOD subsets. We then propose a Dynamic Prototype Refinement strategy (\textit{cf.} Sec.~\ref{sec:dynamic_prototype}), which accumulates selected csID features in class-wise buffers and periodically refines class-specific target prototypes. Lastly, we introduce Contrastive Prototype Entropy Adaptation (\textit{cf.} Sec.~\ref{sec: entropy_contrastive_optimization}), a unified objective that minimizes entropy for csID samples and maximizes entropy for csOOD samples, while further using contrastive learning to increase pseudo-label-relative similarity margins for csID features and reduce the maximum similarity between csOOD features and initialized target prototypes.

\subsection{Uncertainty-Driven ID/OOD Separation}
\label{sec: peparation}

In open-set test-time adaptation, distinguishing between csID and csOOD samples is crucial for ensuring robust adaptation. This is particularly important because minimizing entropy on unknown-class samples can push them toward overconfident known-class predictions and induce harmful updates.
Therefore, we propose the \textbf{Uncertainty-Driven ID/OOD Separation}, which combines prototype similarity with an evidence-derived uncertainty score to distinguish csID from csOOD samples. Specifically, we design an ID/OOD separation score for each sample $x_t$, denoted by $S(x_t)$, based on two criteria: \textbf{1)} the similarity of the sample to known-class classifier prototypes, and \textbf{2)} an evidence-derived uncertainty score computed from the online-model logits.

\textbf{Prototype-Based Similarity.} To estimate whether a target sample is close to the known source classes, we use the class-wise classifier weights as known-class prototypes. Specifically, we maintain a non-optimized reference copy
$M_{\mathrm{ref}}=(g_{\mathrm{ref}},h_{\mathrm{ref}})$ initialized from the source model. The learnable parameters of this reference model are not updated during test-time adaptation, and $g_{\mathrm{ref}}(x_t)$ is used to extract features for prototype-based scoring. Let $w_c$ denote the classifier weight corresponding to class $c$ in the reference prediction head, where $c \in \{1,\ldots,C\}$.
Since each classifier weight encodes a class-specific decision direction, it can be regarded as a prototype of the corresponding known class. We compute the prototype-based score by measuring the maximum cosine similarity between the sample feature and all classifier prototypes:
\begin{equation}
    S_{\mathrm{proto}}(x_t)
=
1-
\max_c
\left(
\frac{
\langle g_{\mathrm{ref}}(x_t),w_c\rangle
}{
\|g_{\mathrm{ref}}(x_t)\|_2\|w_c\|_2
}
\right).
\end{equation}
where $\langle \cdot,\cdot\rangle$ denotes the inner product and $\|\cdot\|_2$ denotes the $L_2$ norm. A lower $S_{\mathrm{proto}}(x_t)$ indicates that the sample feature is close to a known-class classifier prototype, whereas a higher value suggests that the sample is less aligned with known classes and is more likely to be csOOD.

\textbf{Uncertainty Estimation.} Inspired by Evidential Deep Learning~\cite{sensoy2018evidential,ma2025estimating}, we incorporate uncertainty estimation to assess model confidence in its predictions.
Specifically, we estimate evidence-derived uncertainty from the online model logits
$\mathbf{h}=h_{\phi_t}(g_{\theta_t}(x_t))$. We first transform the logits into Dirichlet concentration parameters:
\begin{equation}
\alpha_c(x_t)=\operatorname{softplus}(h_c(x_t))+1,
\qquad
T(x_t)=\sum_{c=1}^{C}\alpha_c(x_t),
\label{equation3}
\end{equation}
where $T(x_t)$ denotes the total Dirichlet strength. The raw evidence-derived uncertainty score is computed as
\begin{equation}
\begin{aligned}
U_{\mathrm{EU}}(x_t)
&=
\frac{C}{T(x_t)+\epsilon},\\
S_{\mathrm{EU}}(x_t)
&=
\frac{
U_{\mathrm{EU}}(x_t)
-
\min_{x\in\mathcal{B}_t}U_{\mathrm{EU}}(x)
}{
\max_{x\in\mathcal{B}_t}U_{\mathrm{EU}}(x)
-
\min_{x\in\mathcal{B}_t}U_{\mathrm{EU}}(x)
+\epsilon
},
\end{aligned}
\label{equation4}
\end{equation}
where $\epsilon>0$ is a small constant for numerical stability. Before score fusion, the raw uncertainty values are min--max normalized
within the current mini-batch to obtain a bounded, batch-relative score
that is more compatible in scale with the classifier-prototype score.
A larger $S_{\mathrm{EU}}(x_t)$ indicates greater relative uncertainty within the current mini-batch.

\textbf{ID/OOD Separation.}  
After computing the classifier-prototype score and the mini-batch-normalized evidence-derived uncertainty score, we compute the joint separation score as
\begin{equation}
S(x_t)
=
\lambda S_{\mathrm{proto}}(x_t)
+
(1-\lambda)S_{\mathrm{EU}}(x_t),
\label{equation5}
\end{equation}
where $\lambda\in[0,1]$ controls the relative contribution of the geometric and the evidence-derived uncertainty score.
Finally, a threshold $\tau$ is applied to the combined score $S(x_t)$ to classify the sample as either csID or csOOD:
\begin{equation}
\label{eqn: cs_ood_spearation}
    x_t \in \begin{cases} 
\text{csID}, & \text{if } S(x_t) < \tau, \\
\text{csOOD}, & \text{if } S(x_t) \ge \tau.
\end{cases}
\end{equation}

\subsection{Dynamic Prototype Refinement Strategy}
\label{sec:dynamic_prototype}

To better track the evolving csID distribution while reducing the influence of csOOD samples, we introduce a \textbf{Dynamic Prototype Refinement (DPR)} strategy. DPR maintains a set of class-specific target prototypes $\{\mu_c\}_{c=1}^{C}$ in the feature space, which provide class-level anchors for the subsequent contrastive prototype adaptation. Unlike the classifier prototypes used for ID/OOD separation, these target prototypes are refined online from reliable csID samples. In our implementation, all target prototypes are initialized as zero vectors and are updated after every $\eta$ effective mini-batches, where an effective mini-batch contains at least one sample identified as csID. We set $\eta=10$ by default.

For each incoming mini-batch, we first identify the csID subset $\mathcal{B}_{t,\mathrm{csID}}$ using the ID/OOD separation strategy (see Sec. \ref{sec: peparation}). For each selected sample $x\in\mathcal{B}_{t,\mathrm{csID}}$, we extract its feature representation from the feature extractor of the current online model before the final prediction head, \textit{i.e.}, $z_x = g_{\theta_t}(x).$
We then assign a pseudo-label according to the current model prediction:
\begin{equation}
    \hat{y}
    =
    \arg\max_{c}
    h_{\varphi_t}^{c}(g_{\theta_t}(x)),
\end{equation}
where $h_{\varphi_t}^{c}(\cdot)$ denotes the logit of class $c$ and $\hat{y}$ is the pseudo-label assigned to sample $x$.
The extracted features are stored in class-wise buffers according to their pseudo-labels. Let $\mathcal{F}_c$ denote the feature buffer for class $c$ within the current update interval. When a sample
$x\in\mathcal{B}_{t,\mathrm{csID}}$ is pseudo-labeled as class $c$, its feature $z_x$ is added to $\mathcal{F}_c$. A prototype-update counter is advanced only when the current mini-batch contributes at least one selected csID feature. After $\eta$ such effective mini-batches, the normalized buffered mean is computed for each non-empty buffer:
\begin{equation}
    \mu_c^{\mathrm{buf}}
    =
    \frac{
    \frac{1}{|\mathcal{F}_c|}
    \sum_{z\in \mathcal{F}_c} z
    }{
    \left\|
    \frac{1}{|\mathcal{F}_c|}
    \sum_{z\in \mathcal{F}_c} z
    \right\|_2
    } .
\end{equation}
The prototype of class $c$ is updated with a momentum rule:
\begin{equation}
    \mu_c
    \leftarrow
    \beta \mu_c
    +
    (1-\beta)\mu_c^{\mathrm{buf}},
    \label{equation9}
\end{equation}
where $\beta\in[0,1]$ is the momentum coefficient. Since prototypes are initialized as zero vectors, if $\mu_c$ has not been initialized by any previous non-empty buffer, we directly set
\begin{equation}
    \mu_c \leftarrow \mu_c^{\mathrm{buf}}.
\end{equation}
If $\mathcal{F}_c$ is empty during the current update interval, $\mu_c$ remains unchanged. After each update, the prototype is normalized as
\begin{equation}
    \mu_c
    \leftarrow
    \frac{\mu_c}{\|\mu_c\|_2}.
    \label{equation11}
\end{equation}
Finally, all class-wise buffers are cleared for the next update interval. 
This effective-batch-based refinement allows the target prototypes to gradually follow the shifted csID distribution while avoiding prototype updates when no reliable csID features are available.  
These refined prototypes serve as reliable anchors for contrastive prototype entropy adaptation.

\subsection{Contrastive Prototype Entropy Adaptation}
\label{sec: entropy_contrastive_optimization}
To ensure effective adaptation on an unlabeled target stream that mixes csID and csOOD samples, we introduce \textbf{Contrastive Prototype Entropy Adaptation (CPEA)}. CPEA combines sample-level entropy optimization with prototype-level contrastive learning to promote confident adaptation on csID samples while suppressing csOOD-induced negative adaptation. The overall objective is defined as
\begin{equation}
\label{eqn: total_loss}
    \mathcal{L}_{t,{\text{total}}} = \mathcal{L}_{\text{entropy}} + \gamma \mathcal{L}_{\text{contrast}},
\end{equation}
where $\gamma\ge0$ is the trade-off hyperparameter.

\textbf{Entropy Optimization.}
Inspired by previous work~\cite{katz2022training,gao2024unified}, we apply different entropy objectives to samples separated by Eq.~(\ref{eqn: cs_ood_spearation}). Specifically, entropy minimization on csID samples encourages confident predictions for shifted known-class samples and supports known-class adaptation. In contrast, entropy maximization on csOOD samples discourages overconfident assignment to any known class, thereby reducing harmful updates induced by unknown-class samples. The entropy losses are defined as
\begin{equation}
    \mathcal{L}_{t,\text{csID}} = \frac{1}{| \mathcal{B}_{t,\text{csID}} |} \sum_{x \in \mathcal{B}_{t,\text{csID}}} H(f_{\theta_t}(x)), 
    \label{equation13}
\end{equation}
\begin{equation}
    \mathcal{L}_{t,\text{csOOD}} = \frac{1}{|\mathcal{B}_{t,\text{csOOD}} |} \sum_{x \in \mathcal{B}_{t,\text{csOOD}}} H(f_{\theta_t}(x)), 
\end{equation}
\begin{equation}
    \label{eqn: entropy}
    \mathcal{L}_{t,\text{entropy}} = \mathcal{L}_{t,\text{csID}} - \alpha_0 \mathcal{L}_{t,\text{csOOD}}-\alpha_1H(\bar{f}_{\theta_t}),
\end{equation}
where $H(f_{\theta_t}(x)) = - \sum_{c=1}^{C} f_{\theta_t}^c(x) \log f_{\theta_t}^c(x)$ is entropy of the softmax output $ f_{\theta_t}(x)$, and $\bar{f}_{\theta_t}=\frac{1}{| \mathcal{B}_t |} \sum_{x \in \mathcal{B}_t} f_{\theta_t}(x)$ is average softmax output over the batch $\mathcal{B}_t$. Here, $\alpha_0$ controls entropy maximization on selected csOOD samples, whereas $\alpha_1$ controls the diversity regularizer on the mean batch prediction, which discourages prediction collapse onto a small subset of known classes.

\textbf{Contrastive Optimization.}
To refine the feature space for csID samples and improve csOOD detection, we employ a contrastive optimization that aligns csID features with their corresponding class prototypes while repelling csOOD features from all prototypes. For samples identified as csID by the ID/OOD separation module, we assign pseudo-labels using the online model prediction and apply an InfoNCE-style loss to pull their normalized features toward the corresponding dynamic prototype:
\begin{equation}
    \mathcal{L}_{\text{InfoNCE}} = -\frac{1}{|\mathcal{B}_{t,\text{csID}}|} \sum_{x \in \mathcal{B}_{t,\text{csID}}} \log \frac{\exp(\cos \theta(z, \mu_{\hat{y}})/\tau_0)}{\sum_{c=1}^C \exp(\cos \theta(z, \mu_c)/\tau_0)},
    \label{InfoNCE}
\end{equation}
where $z$ is the normalized feature, $\mu_{\hat{y}}$ is the prototype for the pseudo-label $\hat{y}$, and $\cos \theta$ is the cosine similarity. For csOOD samples, we introduce a repulsion term that minimizes their maximum cosine similarity to any prototype, effectively pushing their features away from csID class representations: 
\begin{equation}
    \mathcal{L}_{\text{OOD-repulse}} = \frac{1}{|\mathcal{B}_{t,\text{csOOD}}|} \sum_{x \in \mathcal{B}_{t,\text{csOOD}}} \max_c \cos \theta(z, \mu_c).
    \label{OOD-repulse}
\end{equation}
The contrastive loss in Eq. (\ref{eqn: total_loss}) is as follows:
\begin{equation}
\mathcal{L}_{\mathrm{contrast}}
=
\mathcal{L}_{\mathrm{InfoNCE}}
+
\rho\mathcal{L}_{\mathrm{OOD\mbox{-}repulse}},
\end{equation}
where $\rho>0$ controls the contribution of the csOOD repulsion term. We set $\rho=0.1$ in all experiments. When the csID subset is empty, its entropy term and the entire contrastive loss are set to zero, whereas for an empty csOOD subset, only its entropy and OOD-repulsion terms are set to zero, with InfoNCE retained.

\subsection{Theoretical Motivations and Geometric Interpretation}
\label{sec:theoretical_properties}
This subsection provides a theoretical characterization of the three
principal mechanisms underlying \methodname{}. We first establish a sufficient
margin condition for the separability of the fused ID/OOD score and
analyze its stability under bounded score perturbations. We then prove
the one-step angular non-expansion of DPR toward the current buffered
class direction. Finally, through similarity-gradient analysis, we characterize the
stage-dependent prototype-relative effects of CPEA under
zero-initialized target prototypes and establish the full
nearest-prototype max--min geometric equivalence once all target
prototypes have been initialized.

\paragraph{Conditional interpretation of the evidence-derived score}
Recall from Eqs.~\eqref{equation3}--\eqref{equation4} that \methodname{} first
computes the raw evidence-derived uncertainty score
$U_{\mathrm{EU}}(x)=C/(T(x)+\epsilon)$ and then applies mini-batch
min--max normalization. The raw score is strictly decreasing with
respect to the total Dirichlet strength:
\begin{equation}
\frac{\partial U_{\mathrm{EU}}}{\partial T}
=
-\frac{C}{(T+\epsilon)^2}
<0.
\label{eq:eu_monotonicity}
\end{equation}

Consequently, if an ID--OOD pair satisfies
$T(x_{\mathrm{id}})>T(x_{\mathrm{ood}})$, then
$U_{\mathrm{EU}}(x_{\mathrm{id}})
<
U_{\mathrm{EU}}(x_{\mathrm{ood}})$.
In particular, suppose that there exist constants
$m_{\mathrm{id}}$ and $m_{\mathrm{ood}}$ such that
\begin{equation}
T(x_{\mathrm{id}})
\ge
m_{\mathrm{id}}
>
m_{\mathrm{ood}}
\ge
T(x_{\mathrm{ood}})
\label{eq:conditional_strength_gap}
\end{equation}
for all csID and csOOD samples under consideration. This uniform
strength gap is sufficient for the raw score to rank every csID sample
below every csOOD sample. This result is conditional on the learned
logits and does not imply that a source-trained classifier necessarily
satisfies Eq.~\eqref{eq:conditional_strength_gap}.

For a fixed mini-batch, the normalization in
Eq.~\eqref{equation4} applies the same strictly increasing affine
transformation to all raw scores in the batch. It therefore preserves
the ordering of unequal raw scores:
\begin{equation}
U_{\mathrm{EU}}(x_i)
<
U_{\mathrm{EU}}(x_j)
\Longrightarrow
S_{\mathrm{EU}}(x_i)
<
S_{\mathrm{EU}}(x_j).
\label{eq:within_batch_order}
\end{equation}

However, the extrema used in the normalization depend on the current
mini-batch. A raw-score threshold $\tau_{\mathrm{EU}}$ is mapped in
batch $\mathcal{B}_t$ to
\begin{equation}
\widetilde{\tau}_{t}
=
\frac{
\tau_{\mathrm{EU}}
-
\min_{x\in\mathcal{B}_t}U_{\mathrm{EU}}(x)
}{
\max_{x\in\mathcal{B}_t}U_{\mathrm{EU}}(x)
-
\min_{x\in\mathcal{B}_t}U_{\mathrm{EU}}(x)
+
\epsilon
},
\label{eq:batch_dependent_threshold}
\end{equation}
which is generally batch dependent. Moreover, although
$\operatorname{softmax}(\mathbf{h}+a\mathbf{1})
=
\operatorname{softmax}(\mathbf{h})$, the corresponding value of
$U_{\mathrm{EU}}$ generally changes under the transformation
$\mathbf{h}\mapsto\mathbf{h}+a\mathbf{1}$. Therefore,
$U_{\mathrm{EU}}$ should be interpreted as an evidence-derived
uncertainty score that depends on the logit parameterization, rather
than as a calibrated estimate of epistemic uncertainty. The
normalization yields $S_{\mathrm{EU}}\in[0,1)$, whereas
$S_{\mathrm{proto}}\in[0,2]$. Accordingly, $\lambda$ balances two
cues defined over different numerical ranges.
We next characterize the fused score used by~\methodname{} in
Eq.~\eqref{equation5}. For analysis only, let
$\mathcal{I}_t$ and $\mathcal{O}_t$ denote the non-empty latent csID
and csOOD subsets of $\mathcal{B}_t$, respectively. For each pair
$(x_i,x_o)\in\mathcal{I}_t\times\mathcal{O}_t$, define
\begin{align}
\Delta_{\mathrm{proto}}(x_o,x_i)
&=
S_{\mathrm{proto}}(x_o)
-
S_{\mathrm{proto}}(x_i),\\
\Delta_{\mathrm{EU}}(x_o,x_i)
&=
S_{\mathrm{EU}}(x_o)
-
S_{\mathrm{EU}}(x_i),
\end{align}
and define the minimum fused pairwise margin as
\begin{equation}
M_t(\lambda)
=
\min_{\substack{
x_i\in\mathcal{I}_t\\
x_o\in\mathcal{O}_t
}}
\left[
\lambda\Delta_{\mathrm{proto}}(x_o,x_i)
+
(1-\lambda)\Delta_{\mathrm{EU}}(x_o,x_i)
\right].
\label{eq:fused_minimum_margin}
\end{equation}

\noindent\textbf{Proposition 1 (Conditional separability of the fused
score).}
If $M_t(\lambda)>0$, then
\begin{equation}
\max_{x_i\in\mathcal{I}_t}S(x_i)
<
\min_{x_o\in\mathcal{O}_t}S(x_o).
\label{eq:fused_batch_separation}
\end{equation}

Consequently, every threshold
\begin{equation}
\tau_t\in
\left(
\max_{x_i\in\mathcal{I}_t}S(x_i),
\min_{x_o\in\mathcal{O}_t}S(x_o)
\right]
\label{eq:fused_threshold_interval}
\end{equation}
separates the two subsets in batch $t$ according to
Eq.~\eqref{eqn: cs_ood_spearation}.

\emph{Proof.}
By Eqs.~\eqref{equation5} and~\eqref{eq:fused_minimum_margin}, every
$(x_i,x_o)\in\mathcal{I}_t\times\mathcal{O}_t$ satisfies
\begin{equation}
S(x_o)-S(x_i)\ge M_t(\lambda)>0.
\end{equation}

Taking the minimum over $x_o$ and the maximum over $x_i$ yields
Eq.~\eqref{eq:fused_batch_separation}. The decision rule then gives
Eq.~\eqref{eq:fused_threshold_interval}. \hfill$\square$

A single fixed threshold is valid over a stream whenever
\begin{equation}
\sup_t\max_{x_i\in\mathcal{I}_t}S(x_i)
<\tau\le
\inf_t\min_{x_o\in\mathcal{O}_t}S(x_o).
\label{eq:stream_threshold_condition}
\end{equation}
This proposition also makes the role of cue complementarity explicit.
If both pairwise cue margins are positive, every convex combination
preserves the ordering; if one cue is reversed, separation depends on
$\lambda$ and the two margin magnitudes. If the two component scores
are perturbed by at most $\delta_{\mathrm{proto}}$ and
$\delta_{\mathrm{EU}}$ per sample, the ordering remains robust whenever
$M_t(\lambda)>2[\lambda\delta_{\mathrm{proto}}+(1-\lambda)
\delta_{\mathrm{EU}}]$.

Compared with GMM-based gating, the raw evidence-derived score $U_{\mathrm{EU}}(x)$ is computed directly from the sample logits without fitting additional mixture parameters. Its normalized counterpart $S_{\mathrm{EU}}(x)$ remains batch-relative because it depends on the extrema of the current mini-batch. Nevertheless, the normalization does not require an additional distribution-fitting stage. Combined with the classifier-prototype score, it provides predictive and geometric cues for csID/csOOD partitioning in streaming test-time adaptation.

\paragraph{Directional alignment induced by DPR}
Consider class $c$ with an initialized target prototype and a nonzero
buffered feature mean. Let
$\mu_c,\mu_c^{\mathrm{buf}}\in\mathbb{S}^{d-1}$ denote their normalized
directions. Combining Eqs.~\eqref{equation9}--\eqref{equation11} gives
\begin{equation}
v_c
=
\beta\mu_c
+
(1-\beta)\mu_c^{\mathrm{buf}},
\qquad
\mu_c^+
=
\frac{v_c}{\|v_c\|_2},
\qquad
\beta\in[0,1].
\label{eq:dpr_directional_update}
\end{equation}

\noindent\textbf{Proposition 2 (One-step angular non-expansion under
DPR).}
Under Eq.~\eqref{eq:dpr_directional_update}, if
$\mu_c\neq-\mu_c^{\mathrm{buf}}$, then
\begin{equation}
\angle(\mu_c^+,\mu_c^{\mathrm{buf}})
\le
\angle(\mu_c,\mu_c^{\mathrm{buf}}).
\label{eq:dpr_angle_contraction}
\end{equation}

\emph{Proof.}
Let
$\theta=\angle(\mu_c,\mu_c^{\mathrm{buf}})\in[0,\pi)$.
For $\theta=0$, the result is immediate. For
$\theta\in(0,\pi)$, there exists a unit vector $e$ orthogonal to
$\mu_c^{\mathrm{buf}}$ in the two-dimensional subspace spanned by the
two directions such that
\begin{equation}
\mu_c
=
\cos\theta\,\mu_c^{\mathrm{buf}}
+
\sin\theta\,e.
\end{equation}
Substituting this expression into Eq.~\eqref{eq:dpr_directional_update}
gives
\begin{equation}
v_c
=
\left[
\beta\cos\theta+(1-\beta)
\right]\mu_c^{\mathrm{buf}}
+
\beta\sin\theta\,e.
\end{equation}

Let
$\phi=\angle(\mu_c^+,\mu_c^{\mathrm{buf}})$.
The assumption
$\mu_c\neq-\mu_c^{\mathrm{buf}}$ ensures that $v_c\neq 0$, so
$\phi$ is well defined. Direct calculation yields
\begin{equation}
\sin(\theta-\phi)
=
\frac{(1-\beta)\sin\theta}{\|v_c\|_2}
\ge
0.
\end{equation}
Because $\theta-\phi\in(-\pi,\pi)$, the above inequality implies
$\theta-\phi\ge0$, and hence $\phi\le\theta$. This proves
Eq.~\eqref{eq:dpr_angle_contraction}. \hfill$\square$

Proposition 2 establishes one-step alignment toward the buffered class
direction, rather than toward the unobserved target-class center.
Therefore, interpreting DPR as tracking the target distribution requires
that the ID/OOD separation and pseudo-label assignments yield a buffered
direction $\mu_c^{\mathrm{buf}}$ that sufficiently approximates the csID
class direction.

\paragraph{Prototype-relative geometry induced by CPEA}
The entropy terms in
Eqs.~\eqref{equation13}--\eqref{eqn: entropy} operate on output
probabilities and contain no explicit feature--prototype distance.
Their prototype-relative geometric effect is therefore indirect.
At adaptation step $t$, let
\[
\mathcal{V}_t
=
\left\{
c\in\{1,\ldots,C\}:
\|\mu_c\|_2>0
\right\}
\]
denote the set of initialized target prototypes. In the implementation,
the contrastive terms are evaluated over all $C$ prototype coordinates.
Under the epsilon-stabilized row-wise normalization used in the
implementation, every nonzero prototype is normalized whereas each zero
prototype remains unchanged. Therefore, $\|\mu_c\|_2=1$ for
$c\in\mathcal{V}_t$, whereas $\mu_c=\mathbf{0}$ for
$c\notin\mathcal{V}_t$. Let $z\in\mathbb{S}^{d-1}$ and
$s_c=\langle z,\mu_c\rangle$ for $c\in\{1,\ldots,C\}$.

For a selected csID sample with pseudo-label $\hat y$, define
\begin{equation}
q_c
=
\frac{\exp(s_c/\tau_0)}
{\sum_{j=1}^{C}\exp(s_j/\tau_0)}.
\label{eq:theory_proto_prob}
\end{equation}
The single-sample InfoNCE term satisfies
\begin{equation}
\frac{\partial \ell_{\mathrm{InfoNCE}}}{\partial s_{\hat y}}
=
\frac{q_{\hat y}-1}{\tau_0}
\leq 0,
\qquad
\frac{\partial \ell_{\mathrm{InfoNCE}}}{\partial s_c}
=
\frac{q_c}{\tau_0}
\geq 0,
\quad c\neq\hat y,
\label{eq:theory_infonce_gradients}
\end{equation}
and the ambient Euclidean gradient with respect to the normalized
coordinate $z$ is
\begin{equation}
\nabla_z\ell_{\mathrm{InfoNCE}}
=
\frac{1}{\tau_0}
\left(
\sum_{c=1}^{C}q_c\mu_c-\mu_{\hat y}
\right).
\label{eq:theory_infonce_feature_gradient}
\end{equation}

When $\hat y\in\mathcal{V}_t$, minimizing the InfoNCE term favors a
larger relative margin between the initialized pseudo-label prototype
and its competitors. Since, for each $c\in\mathcal{V}_t$,
\begin{equation}
\|z-\mu_c\|_2^2
=
2-2\langle z,\mu_c\rangle,
\label{eq:hypersphere_identity}
\end{equation}
increasing $s_{\hat y}$ reduces the corresponding Euclidean and angular
distances on the unit sphere. With correct pseudo-labels and informative
prototypes, this provides a prototype-centered regularization effect;
it does not by itself guarantee global intra-class compactness. When
$\hat y\notin\mathcal{V}_t$, the pseudo-label prototype remains zero,
so the loss provides no attraction toward that uninitialized prototype.
If $\mathcal{V}_t=\varnothing$, all similarities are zero, and the
InfoNCE loss equals $\log C$ with zero gradient with respect to $z$.
For a selected csOOD sample in a mini-batch where the contrastive loss is active, the implemented repulsion term is
\begin{equation}
\ell_{\mathrm{rep}}(z)
=
\max_{1\leq c\leq C}
\langle z,\mu_c\rangle.
\label{eq:implemented_repulsion}
\end{equation}
When all target prototypes are initialized, i.e.,
$\mathcal{V}_t=\{1,\ldots,C\}$,
Eq.~\eqref{eq:hypersphere_identity} gives
\begin{equation}
\arg\min_{z\in\mathbb{S}^{d-1}}
\max_{1\leq c\leq C}\langle z,\mu_c\rangle
=
\arg\max_{z\in\mathbb{S}^{d-1}}
\min_{1\leq c\leq C}\|z-\mu_c\|_2^2.
\label{eq:repulsion_maxmin}
\end{equation}
Thus, in the fully initialized regime, minimizing the repulsion term is
equivalent to maximizing the minimum squared distance between a selected
csOOD feature and the target-prototype set.

If $\mathcal{V}_t=\varnothing$, all target prototypes are zero and
$\ell_{\mathrm{rep}}(z)=0$. When
$\emptyset\neq\mathcal{V}_t\subsetneq\{1,\ldots,C\}$,
at least one target prototype remains uninitialized. The corresponding
zero similarities yield
\begin{equation}
\ell_{\mathrm{rep}}(z)
=
\max
\left\{
0,\,
\max_{c\in\mathcal{V}_t}
\langle z,\mu_c\rangle
\right\}.
\label{eq:partial_repulsion}
\end{equation}
The repulsion term therefore penalizes only the maximum positive
similarity between the selected csOOD feature and the initialized
prototypes. Together with InfoNCE, it provides stage-dependent,
prototype-relative regularization for csID alignment and csOOD
separation.

\begin{table*}[t]
\renewcommand{\arraystretch}{1.0}
\renewcommand{\tabcolsep}{3pt}
\begin{center}
\caption{Performance comparison on CIFAR-10-C and CIFAR-100-C using a WideResNet-40-2 backbone. $\uparrow$ indicates higher is better, $\downarrow$ lower is better. Best in \textbf{bold} and second best in \underline{underlined}. `\textbf{Ours}' denotes TENT with our \methodname.}
\label{tab:cifar_results}
\begin{tabular}{lcccccccccccc}
\toprule
\multirow{2}{*}{\textbf{Method}}
  & \multicolumn{4}{c}{\textbf{CIFAR-10-C}}
  & \multicolumn{4}{c}{\textbf{CIFAR-100-C}}
  & \multicolumn{4}{c}{\textbf{Average}} \\
\cmidrule(lr){2-5} \cmidrule(lr){6-9} \cmidrule(lr){10-13}
  & {\makecell[c]{Acc$\uparrow$}}
  & {\makecell[c]{AUROC$\uparrow$}}
  & {\makecell[c]{FPR@95TPR$\downarrow$}}
  & {\makecell[c]{OSCR$\uparrow$}}
  & {\makecell[c]{Acc$\uparrow$}}
  & {\makecell[c]{AUROC$\uparrow$}}
  & {\makecell[c]{FPR@95TPR$\downarrow$}}
  & {\makecell[c]{OSCR$\uparrow$}}
  & {\makecell[c]{Acc$\uparrow$}}
  & {\makecell[c]{AUROC$\uparrow$}}
  & {\makecell[c]{FPR@95TPR$\downarrow$}}
  & {\makecell[c]{OSCR$\uparrow$}} \\
\midrule
Source            & 81.73 & 77.89 & 79.47 & 68.44 & 53.25 & 60.55 & 94.97 & 39.87 & 67.49 & 69.22 & 87.22 & 54.16 \\
BN Adapt~\cite{li2017revisiting}          & 84.20 & 80.40 & 76.85 & 72.13 & 57.16 & 72.45 & 84.29 & 47.10 & 70.68 & 76.43 & 80.57 & 59.62 \\
CoTTA~\cite{wang2022cotta}             & \textbf{85.74} & 85.81 & 72.77 & 77.17 & 56.50 & 77.06 & 80.98 & 49.00 & 71.12 & 81.44 & 76.88 & 63.09 \\
\midrule
TENT~\cite{wang2021tent}             & 79.12 & 67.79 & 95.76 & 58.27 & 50.19 & 59.42 & 93.54 & 36.25 & 64.66 & 63.61 & 94.65 & 47.26 \\
EATA~\cite{niu2022eata}            & 76.17 & 55.43 & 92.88 & 46.66 & \textbf{61.34} & 87.05 & \underline{56.06} & \underline{57.27} & 68.76 & 71.24 & 74.47 & 51.97 \\
SAR~\cite{niutowards}             & 78.41 & 66.17 & 96.13 & 56.79 & 59.67 & 71.70 & 83.56 & 48.68 & 69.04 & 68.94 & 89.85 & 52.74 \\
COME~\cite{zhangcome}              & \underline{85.31} & 75.34 & 84.01 & 69.29 & 58.09 & 72.28 & 79.24 & 47.49 & 71.70 & 73.81 & 81.63 & 58.39 \\
OSTTA~\cite{lee2023towards}            & 84.24 & 72.67 & 76.98 & 64.98 & 60.03 & 75.09 & 82.78 & 51.23 & 72.14 & 73.88 & 79.88 & 58.11 \\
UniEnt~\cite{gao2024unified}       & 84.34 & \underline{92.15} & \underline{37.25} & \underline{80.24} & 59.01 & \underline{88.07} & 57.17 & 55.95 & 71.68 & \underline{90.11} & \underline{47.21} & \underline{68.10} \\
ROSETTA~\cite{zhao2026mitigating}       & 84.90 & 85.06 & 66.00 & 76.34 & 59.53 & 87.02 & 61.61 & 56.07 & \underline{72.22} & 86.04 & 63.81 & 66.21 \\
\rowcolor{pink!30}\textbf{\methodname{} (Ours)}
                   & 84.53 & \textbf{96.18} & \textbf{13.57} & \textbf{82.27}
                   & \underline{60.14} & \textbf{93.26} & \textbf{27.59} & \textbf{58.08}
                   & \textbf{72.34} & \textbf{94.72} & \textbf{20.58} & \textbf{70.18} \\
\bottomrule
\end{tabular}
\end{center}
\end{table*}

\begin{table*}[t]
\renewcommand{\arraystretch}{1.0}
\renewcommand{\tabcolsep}{3pt}
\begin{center}
\caption{Performance comparison on Tiny-ImageNet-C across different backbones (ResNet-50 and ViT-Base). `-' indicates the experiment was not applicable. Best in \textbf{bold} and second best in \underline{underlined}. `\textbf{Ours}' is TENT with our \methodname.}
\label{tab:tiny_imagenet_results}
\begin{tabular}{lcccccccccccc}
\toprule
\multirow{2}{*}{\textbf{Method}}
  & \multicolumn{4}{c}{\textbf{ResNet-50}}
  & \multicolumn{4}{c}{\textbf{ViT-Base}}
  & \multicolumn{4}{c}{\textbf{Average}} \\
\cmidrule(lr){2-5} \cmidrule(lr){6-9} \cmidrule(lr){10-13}
  & {\makecell[c]{Acc$\uparrow$}}
  & {\makecell[c]{AUROC$\uparrow$}}
  & {\makecell[c]{FPR@95TPR$\downarrow$}} 
  & {\makecell[c]{OSCR$\uparrow$}}
  & {\makecell[c]{Acc$\uparrow$}}
  & {\makecell[c]{AUROC$\uparrow$}}
  & {\makecell[c]{FPR@95TPR$\downarrow$}}
  & {\makecell[c]{OSCR$\uparrow$}}
  & {\makecell[c]{Acc$\uparrow$}}
  & {\makecell[c]{AUROC$\uparrow$}}
  & {\makecell[c]{FPR@95TPR$\downarrow$}}
  & {\makecell[c]{OSCR$\uparrow$}} \\
\midrule
Source      & 22.36 & 48.11 & 90.80 & 15.81 & 22.09 & 48.45 & 94.85 & 16.29 & 22.23 & 48.28 & 92.83 & 16.05 \\
BN Adapt~\cite{li2017revisiting}    & 33.79 & 56.43 & 87.80 & 25.60 & - & - & - & - & - & - & - & - \\
CoTTA~\cite{wang2022cotta}        & 32.68 & 47.60 & 92.27 & 22.45 & 16.17 & 40.99 & 95.38 & 10.30 & 24.43 & 44.30 & 93.83 & 16.38 \\
\midrule
TENT~\cite{wang2021tent}        & 25.45 & 52.22 & 94.29 & 17.48 & 26.45 & \underline{52.70} & 95.13 & 19.65 & 25.95 & 52.46 & 94.71 & 18.57 \\
EATA~\cite{niu2022eata}        & \underline{34.46} & 55.86 & 92.52 & 25.92 & \underline{27.95} & 52.61 & 95.14 & \underline{21.31} & \textbf{31.21} & 54.24 & 93.83 & \underline{23.62} \\
SAR~\cite{niutowards}         & 34.41 & 54.18 & 93.74 & 24.99 & 22.34 & 48.64 & \underline{94.81} & 16.37 & 28.38 & 51.41 & 94.28 & 20.68 \\
COME~\cite{zhangcome}        & 34.28 & 49.27 & 93.92 & 23.65 & 22.40 & 48.80 & \textbf{94.70} & 16.41 & 28.34 & 49.04 & 94.31 & 20.03 \\
OSTTA~\cite{lee2023towards}       & \textbf{34.75} & 53.00 & 95.29 & 25.54 & 27.02 & 49.25 & 96.06 & 20.17 & 30.89 & 51.13 & 95.68 & 22.86 \\
UniEnt~\cite{gao2024unified} & 34.45 & 61.77 & 91.53 &  \underline{27.75} & 24.24 & 46.86 & 96.11 & 17.67 & 29.35 & 54.32 & 93.82 & 22.71 \\
ROSETTA~\cite{zhao2026mitigating}       & 34.09 & \underline{65.62} & \underline{87.25} & \underline{27.75} & 23.85 & 48.78 & 96.43 & 16.84 & 28.97 & \underline{57.20} & \underline{91.84} & 22.30 \\
\rowcolor{pink!30}\textbf{\methodname{} (Ours)}
             & 33.85 & \textbf{67.47} & \textbf{81.06} &\textbf{28.10}
             & \textbf{28.53} & \textbf{52.87} & 95.37 & \textbf{21.70}
             & \underline{31.19} & \textbf{60.17} & \textbf{88.22} & \textbf{24.90} \\
\bottomrule
\end{tabular}
\end{center}
\end{table*}

\section{Experiments}
\label{sec: experiments}

\subsection{Experimental Settings}
\textbf{Datasets.} Our experimental setup is designed to construct open-set test streams containing both csID and csOOD samples, where the test data is a composite of both csID and csOOD samples. The csID component is sourced from three standard corruption benchmarks: CIFAR-10-C, CIFAR-100-C, and Tiny-ImageNet-C~\cite{hendrycks2019robustness}, where we consistently use the most severe corruption setting, level 5. To create the matching csOOD component, we adopt the protocol from OSTTA~\cite{lee2023towards} and apply identical corruption functions to the original SVHN and ImageNet-O test sets, generating SVHN-C and ImageNet-O-C. The SVHN-C provides the csOOD samples for the CIFAR-C benchmarks, while ImageNet-O-C is paired with Tiny-ImageNet-C, thus forming the final mixed-distribution test sets. In addition, to evaluate robustness under small-gap settings, we follow the STAMP \cite{yu2024stamp} and construct CIFAR-10/CIFAR-100 splits with an 8:2 ratio of ID to OOD classes.

\textbf{Evaluation protocols.} Following previous studies~\cite{lee2023towards}, the model adapts to sequentially arriving domains without parameter resets. For each mini-batch, predictions are made before the model is updated, ensuring no future information influences current outputs.  In the main evaluation, each mini-batch contains equal numbers of csID and csOOD samples. Additional experiments evaluate varying csOOD ratios from 10\% to 90\%.
Performance is quantified using four metrics. Accuracy (Acc) measures classification performance on csID samples. Open-set detection capability is evaluated via AUROC and FPR@95TPR. To assess the trade-off between classification and detection, we report the Open-Set Classification Rate (OSCR) for a balanced view.

\textbf{Baselines.} We compare \methodname{} with representative closed-set TTA methods, including BN Adapt~\cite{li2017revisiting}, CoTTA~\cite{wang2022cotta}, TENT~\cite{wang2021tent}, EATA~\cite{niu2022eata}, SAR~\cite{niutowards}, and COME~\cite{zhangcome}, as well as open-set TTA methods, including OSTTA~\cite{lee2023towards}, UniEnt \cite{gao2024unified} and ROSETTA~\cite{zhao2026mitigating}.

\textbf{Implementation details.}
Following previous studies~\cite{lee2023towards}, experiments on CIFAR benchmarks adopt WideResNet-40-2~\cite{zagoruyko2016wide} pretrained from RobustBench~\cite{croce2021robustbench}. For Tiny-ImageNet-C, we evaluate on two representative architectures, ResNet-50~\cite{he2016deep} and ViT-Base~\cite{dosovitskiyimage}, to verify architectural generality. ResNet-50 is trained on Tiny-ImageNet using SGD for 50 epochs, while ViT-Base is fine-tuned on the same dataset using Adam for 100 epochs. During TTA, all methods use the Adam optimizer with batch size 100. For CNN-based models, adaptation is applied to Batch Normalization affine parameters, while for ViT-based models we adapt Layer Normalization affine parameters, following standard practice in TTA.
Learning rates for all baselines follow their official implementations or standard settings, including 0.001 for TENT, EATA, and OSTTA; 0.01 for CoTTA; and 0.0001 for SAR and COME. For \methodname{}, we use a learning rate of 0.001. 
The prototype momentum, contrastive temperature, OOD-repulsion coefficient, and prototype-update interval are fixed as
$\beta=0.9$, $\tau_0=0.1$, $\rho=0.1$, and $\eta=10$, respectively.
The remaining trade-off parameters $(\lambda, \tau, \gamma, \alpha_0, \alpha_1)$ are kept fixed within each benchmark. The configurations on Tiny-ImageNet-C with ResNet and ViT: $(0.9, 0.65, 0.2, 1.0, 0.1)$ and $(0.7, 0.9, 0.1, 1.0, 1.0)$. Experiments are conducted on an 80G memory-sized NVIDIA A800 GPU using PyTorch 1.12.1. We use a fixed random seed of 1 for all experiments. Unless otherwise specified, \methodname{} is instantiated with TENT as the base adaptation method in the main experiments.

\begin{table*}[t]
\renewcommand{\arraystretch}{1.0}
\renewcommand{\tabcolsep}{4pt}
\begin{center}
\caption{Ablation experiments on CIFAR-10-C and CIFAR-100-C using a WideResNet-40-2 backbone. `PS' denotes Prototype-based Similarity, `UE' denotes Uncertainty Estimation, `DPR' denotes Dynamic Prototype Refinement strategy, and `CL' denotes Contrastive Learning Optimization.}
\label{tab:ablation_results}
\begin{tabular}{lcccccccccccc}
\toprule
\multirow{2}{*}{\textbf{Method}}
  & \multicolumn{4}{c}{\textbf{CIFAR-10-C}}
  & \multicolumn{4}{c}{\textbf{CIFAR-100-C}}
  & \multicolumn{4}{c}{\textbf{Average}} \\
\cmidrule(lr){2-5} \cmidrule(lr){6-9} \cmidrule(lr){10-13}
  & {\makecell[c]{Acc$\uparrow$}}
  & {\makecell[c]{AUROC$\uparrow$}}
  & {\makecell[c]{FPR@95TPR$\downarrow$}}
  & {\makecell[c]{OSCR$\uparrow$}}
  & {\makecell[c]{Acc$\uparrow$}}
  & {\makecell[c]{AUROC$\uparrow$}}
  & {\makecell[c]{FPR@95TPR$\downarrow$}}
  & {\makecell[c]{OSCR$\uparrow$}}
  & {\makecell[c]{Acc$\uparrow$}}
  & {\makecell[c]{AUROC$\uparrow$}}
  & {\makecell[c]{FPR@95TPR$\downarrow$}}
  & {\makecell[c]{OSCR$\uparrow$}} \\
\midrule
TENT & 79.12 & 67.79 & 95.76  & 58.27 & 50.19 & 59.42 & 93.54  & 36.25 & 64.66 & 63.61 & 94.65  & 47.26  \\
w/o PS & 78.46 & 85.40 & 36.21  & 69.22  & 53.60 & 90.14& \underline{27.18}& 50.16 & 66.03  & 87.77  & \underline{31.70} & 59.69  \\
w/o UE & \textbf{84.87}  & 89.88 & 45.13& 79.10 & 56.96 &\textbf{94.33} & \textbf{26.07}& 55.51& 70.92 & \underline{92.11} & 35.60  & 67.31 \\
w/o DPR & 84.35 & \underline{93.57} & 28.10& \underline{80.95}& 59.64& 86.75& 51.70 & 55.32& 72.00 & 90.16  & 39.90   & 68.14 \\
w/o CL & 84.38 & 93.28  & \underline{26.05} & 80.66 & \underline{60.03}& 89.58 & 39.40& \underline{56.59} & \underline{72.21} & 91.43  & 32.73 & \underline{68.63} \\
\rowcolor{pink!30}\textbf{Ours}
                   & \underline{84.53} & \textbf{96.18} & \textbf{13.57} & \textbf{82.27}
                   & \textbf{60.14} & \underline{93.26} & 27.59 & \textbf{58.08}
                   & \textbf{72.34} & \textbf{94.72} & \textbf{20.58} & \textbf{70.18} \\
\bottomrule
\end{tabular}
\end{center}
\end{table*}

\subsection{Comparison Experiments}

\textbf{CIFAR benchmarks.} As shown in Table \ref{tab:cifar_results}, \methodname{} consistently improves open-set detection performance over strong baselines on CIFAR-10-C and CIFAR-100-C. On CIFAR-10-C, \methodname{} achieves competitive classification accuracy (84.53\%) while obtaining the best AUROC (96.18\%) and the lowest FPR@95TPR (13.57\%). Compared to UniEnt, this corresponds to a 63.6\% relative reduction in false positives, indicating improved separation between csID and csOOD samples. On CIFAR-100-C, \methodname{} further achieves the best performance across all OOD detection metrics, reducing FPR@95TPR to 27.59\%, which is 50.8\% lower than EATA. These results demonstrate consistent improvements in open-set robustness under severe distribution shifts.

\textbf{Tiny-ImageNet-C.} As shown in Table \ref{tab:tiny_imagenet_results}, \methodname{} achieves strong performance across both ResNet-50 and ViT-Base backbones. On ResNet-50, it attains the highest AUROC and OSCR, while maintaining competitive accuracy. On ViT-Base, \methodname{} achieves the best performance in accuracy and OSCR, with an OSCR of 21.70\%, outperforming all compared methods. Overall, the results verify the effectiveness and generality of \methodname{} across architectures.

\subsection{Ablation Studies}
\textbf{Effectiveness of Uncertainty-Driven ID/OOD Separation.} 
We evaluate the contribution of the two components in the proposed ID/OOD separation module by removing Uncertainty Estimation (w/o UE) and Prototype-based Similarity (w/o PS), respectively. As shown in Table~\ref{tab:ablation_results}, removing either component degrades the overall performance. Without Uncertainty Estimation, the average AUROC decreases from 94.72\% to 92.11\%, and the average FPR@95TPR increases from 20.58\% to 35.60\%, indicating that evidence-derived uncertainty is important for identifying uncertain csOOD samples. Removing Prototype-Based Similarity causes a larger drop in OSCR, from 70.18\% to 59.69\%, suggesting that classifier-prototype similarity provides complementary geometric cues for separating csID and csOOD samples.

\textbf{Effectiveness of Dynamic Prototype Refinement Strategy.}
We further remove the DPR strategy (w/o DPR) to evaluate the role of online prototype updates. Compared with the complete method, w/o DPR reduces the average AUROC from 94.72\% to 90.16\% and increases FPR@95TPR from 20.58\% to 39.90\%. This indicates that periodically refining target prototypes with selected csID samples helps maintain reliable class anchors under distribution shifts and reduces prototype contamination from csOOD samples.

\begin{table}[t]
\renewcommand{\arraystretch}{1.0}
\renewcommand{\tabcolsep}{7pt}
\begin{center}
\caption{Results under the small-gap classes setting, where CIFAR-10/CIFAR-100 classes are split into 8:2 (ID: OOD). This setting poses a more challenging scenario.}
\label{tab:small_gap}
\begin{tabular}{lcccc}
\toprule
\textbf{Method}              & {\makecell[c]{Acc$\uparrow$}} & {\makecell[c]{AUROC$\uparrow$}} & {\makecell[c]{FPR@95TPR$\downarrow$}}& {\makecell[c]{OSCR$\uparrow$}} \\ \midrule
Source              & 66.81 & 46.75& 93.99& 36.13 \\
BN Adapt~\cite{li2017revisiting} &72.70&62.06&91.30&52.15 \\
CoTTA~\cite{wang2022cotta} &72.74&61.65&89.96&51.42\\
\midrule
TENT~\cite{wang2021tent}              & 73.68 & 61.16& 94.77 & 51.55 \\
EATA~\cite{niu2022eata}        & 73.16 & 61.34 & 87.49 & 50.52 \\ 
SAR~\cite{niutowards}             & 73.53 & 62.01 & 91.86    & 52.63 \\
COME~\cite{zhangcome}   & 72.62 & \underline{62.92} & 90.86    & \underline{52.81} \\
OSTTA~\cite{lee2023towards} & \underline{73.75} & 59.40   & \underline{86.88}    & 50.32 \\
UniEnt~\cite{gao2024unified} & 73.70 & 62.21   & \textbf{86.69}    & 52.68 \\
ROSETTA~\cite{zhao2026mitigating} & 71.78 & 61.44   & 93.00    & 51.96 \\
\rowcolor{pink!30}\textbf{\methodname{} (Ours)}       & \textbf{74.47} & \textbf{64.30} & 88.03 & \textbf{54.64} \\
\bottomrule
\end{tabular}
\end{center}
\end{table}

\textbf{Effectiveness of Contrastive Prototype Entropy Adaptation.} 
We evaluate the contribution of contrastive optimization by removing the contrastive learning term (w/o CL). From Table~\ref{tab:ablation_results}, removing this term reduces the average OSCR from 70.18\% to 68.63\% and increases FPR@95TPR from 20.58\% to 32.73\%. These results suggest that prototype-level contrastive learning improves feature separation by pulling csID features toward class prototypes while reducing the similarity between csOOD features and known-class prototypes.

\begin{table*}[t]
\renewcommand{\arraystretch}{1.0}
\setlength{\tabcolsep}{3.0pt}
\begin{center}
\caption{Performance comparison of different ID/OOD separation strategies on CIFAR-10-C and CIFAR-100-C using a WideResNet-40-2 backbone. ``Ours (w/ GMM)'' replaces the proposed Uncertainty-Driven ID/OOD Separation with a UniEnt-style GMM gate, while keeping all other components unchanged.}
\label{tab:gmm_result}
\begin{tabular}{lcccccccccccc}
\toprule
\multirow{2}{*}{\textbf{Method}}
  & \multicolumn{4}{c}{\textbf{CIFAR-10-C}}
  & \multicolumn{4}{c}{\textbf{CIFAR-100-C}}
  & \multicolumn{4}{c}{\textbf{Average}} \\
\cmidrule(lr){2-5} \cmidrule(lr){6-9} \cmidrule(lr){10-13}
  & {\makecell[c]{Acc$\uparrow$}}
  & {\makecell[c]{AUROC$\uparrow$}}
  & {\makecell[c]{FPR@95TPR$\downarrow$}}
  & {\makecell[c]{OSCR$\uparrow$}}
  & {\makecell[c]{Acc$\uparrow$}}
  & {\makecell[c]{AUROC$\uparrow$}}
  & {\makecell[c]{FPR@95TPR$\downarrow$}}
  & {\makecell[c]{OSCR$\uparrow$}}
  & {\makecell[c]{Acc$\uparrow$}}
  & {\makecell[c]{AUROC$\uparrow$}}
  & {\makecell[c]{FPR@95TPR$\downarrow$}}
  & {\makecell[c]{OSCR$\uparrow$}} \\
\midrule
Source
& 81.73 & 77.89 & 79.47 & 68.44
& 53.25 & 60.55 & 94.97 & 39.87
& 67.49 & 69.22 & 87.22 & 54.16 \\

\midrule
TENT~\cite{wang2021tent}
& 79.12 & 67.79 & 95.76 & 58.27
& 50.19 & 59.42 & 93.54 & 36.25
& 64.66 & 63.61 & 94.65 & 47.26 \\

Ours (w/ GMM~\cite{gao2024unified})
& \underline{84.45} & \underline{95.67} & \underline{18.16} & \underline{82.12}
& \underline{59.20} & \underline{92.30} & \underline{33.64} & \underline{56.72}
& \underline{71.83} & \underline{93.99} & \underline{25.90} & \underline{69.42} \\

\rowcolor{pink!30}\textbf{\methodname{} (Ours)}
& \textbf{84.53} & \textbf{96.18} & \textbf{13.57} & \textbf{82.27}
& \textbf{60.14} & \textbf{93.26} & \textbf{27.59} & \textbf{58.08}
& \textbf{72.34} & \textbf{94.72} & \textbf{20.58} & \textbf{70.18} \\
\bottomrule
\end{tabular}
\end{center}
\end{table*}

\begin{figure}[t]
\centering
\vspace{-10pt}
\subfloat[Source\label{fig:tsne_source}]{%
\includegraphics[width=0.23\textwidth]{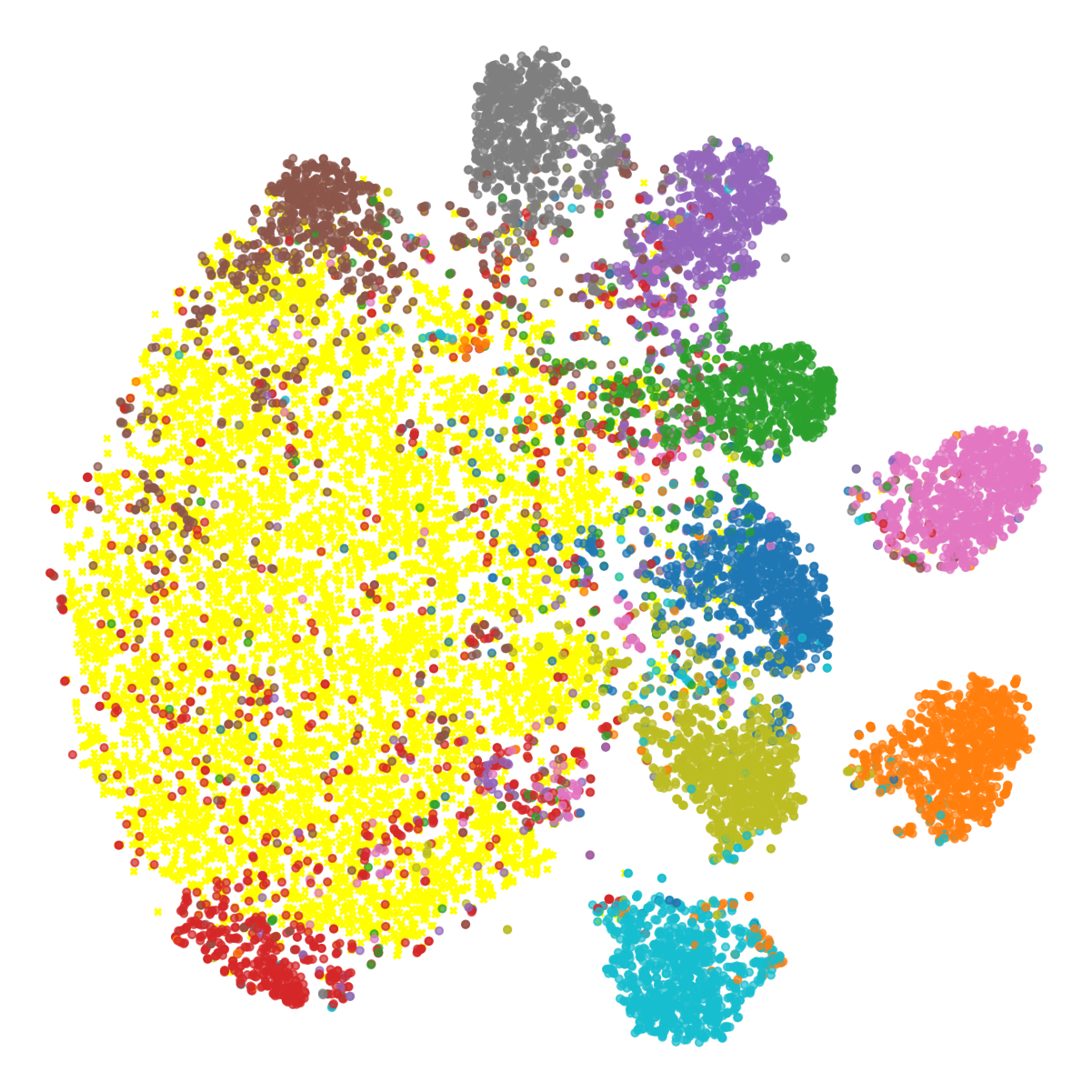}
}\hfill
\subfloat[TENT\label{fig:tsne_tent}]{%
\includegraphics[width=0.23\textwidth]{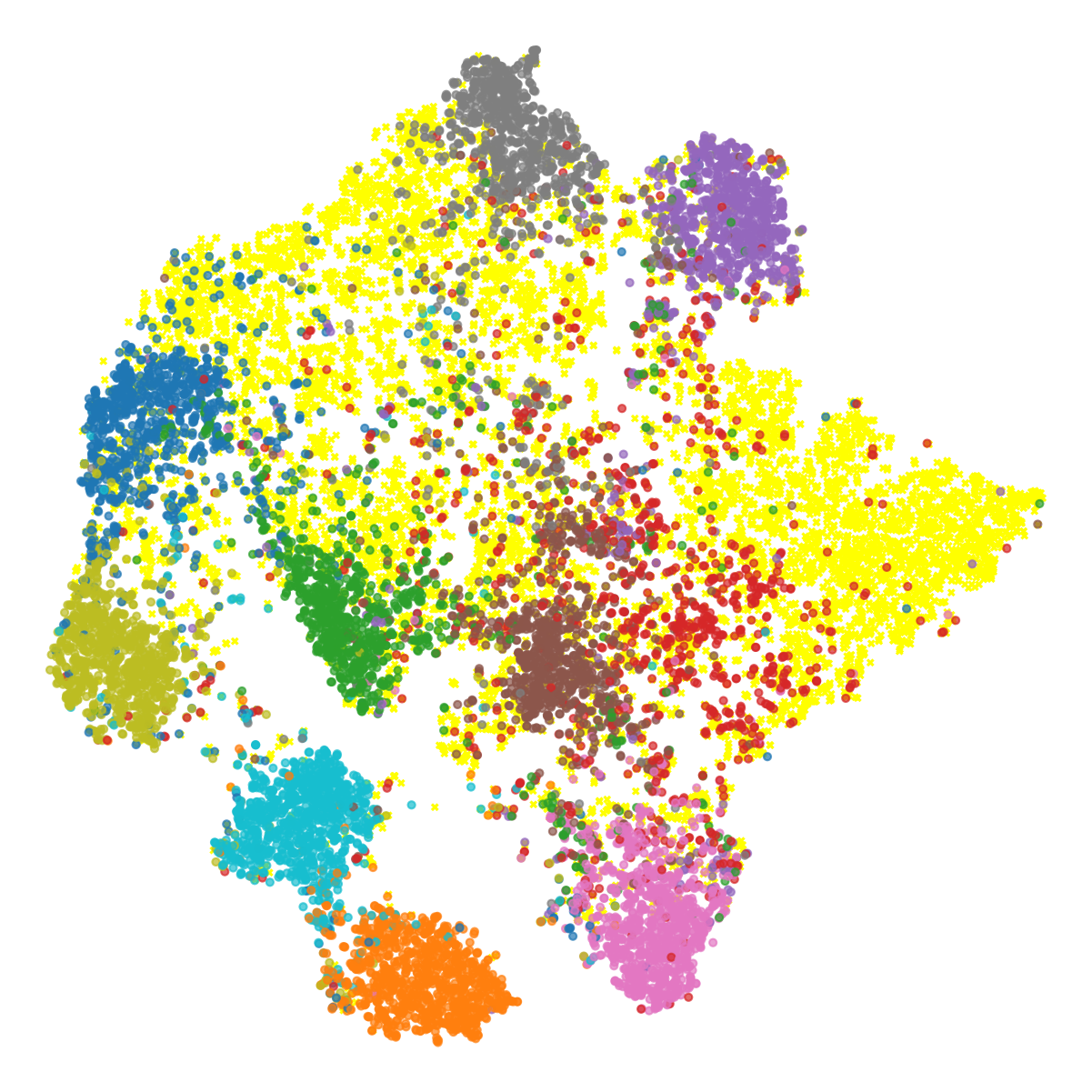}
}\hfill
\subfloat[UniEnt\label{fig:tsne_unient}]{%
\includegraphics[width=0.23\textwidth]{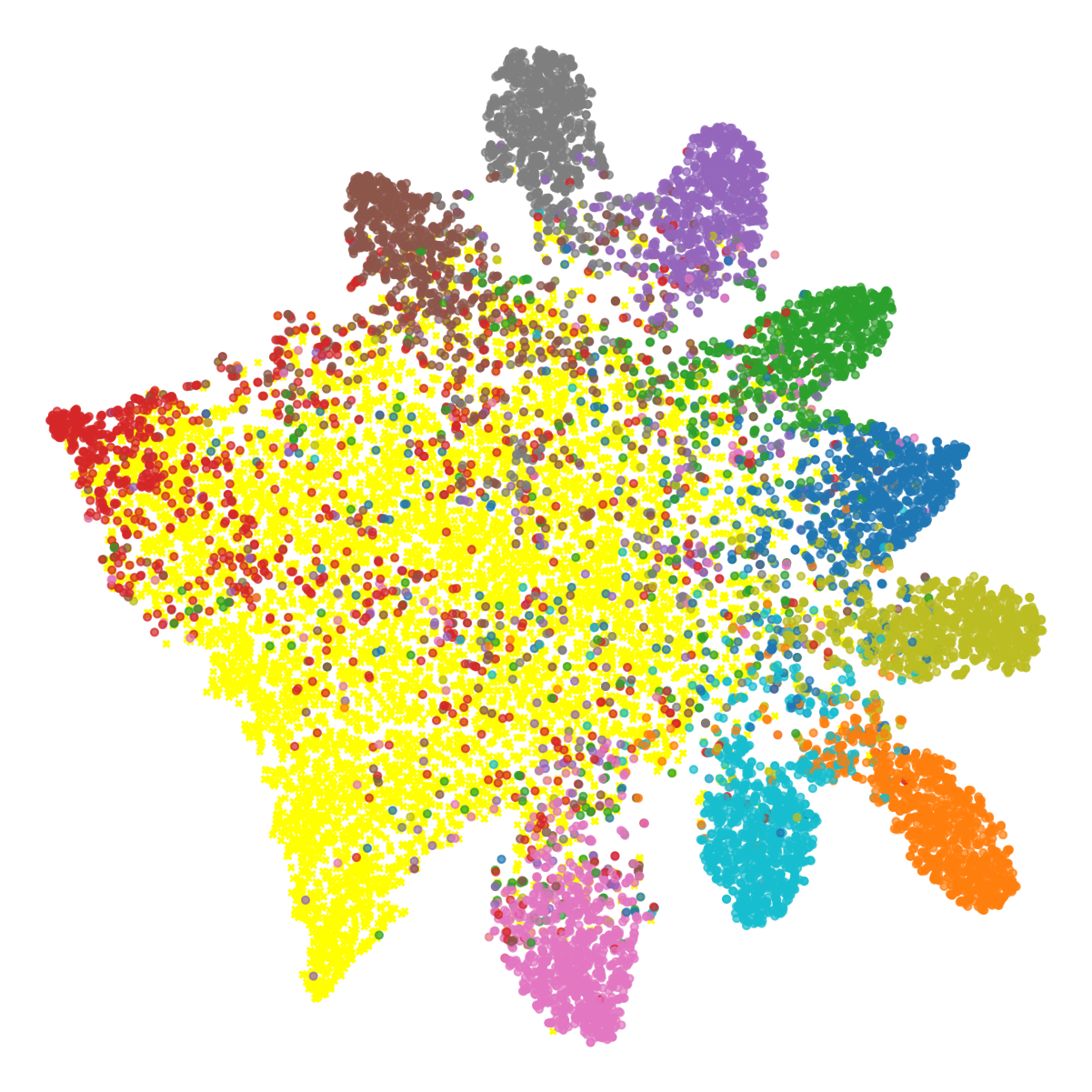}
}\hfill
\subfloat[\methodname{} (Ours)\label{fig:ours}]{%
\includegraphics[width=0.23\textwidth]{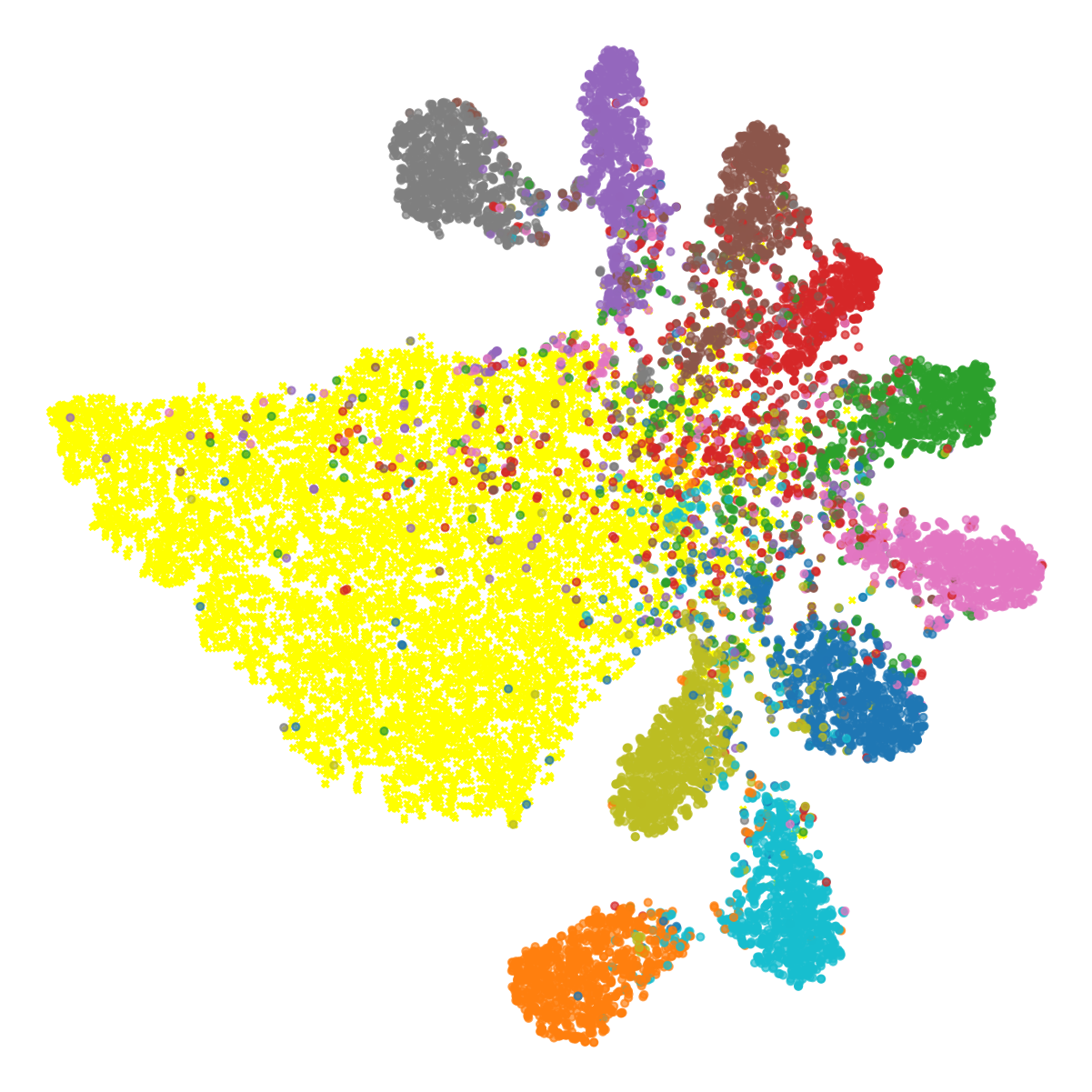}
}
\caption{t-SNE visualization of CIFAR-10-C and SVHN-C features, where SVHN-C samples are treated as csOOD. Yellow denotes csOOD samples, while the remaining colors denote different csID classes.}
\label{fig:tsne}
\end{figure}

\subsection{More Discussions}

\textbf{Robustness under small-gap settings.} 
We evaluate \methodname{} under the small-gap setting, where ID and OOD classes are semantically closer and thus harder to distinguish. As shown in Table~\ref{tab:small_gap}, \methodname{} achieves the best Acc, AUROC, and OSCR among the compared methods, while remaining competitive in FPR@95TPR. Specifically, \methodname{} improves AUROC by 1.38 percentage points and OSCR by 1.83 percentage points over the second-best methods. These results indicate that \methodname{} remains effective when ID and OOD classes are difficult to separate. The improvement is consistent with the design of uncertainty-driven csID/csOOD separation and contrastive prototype adaptation, which provide complementary cues for handling subtle open-set differences.

\begin{figure}[t]
    \centering
    \includegraphics[width=\linewidth]{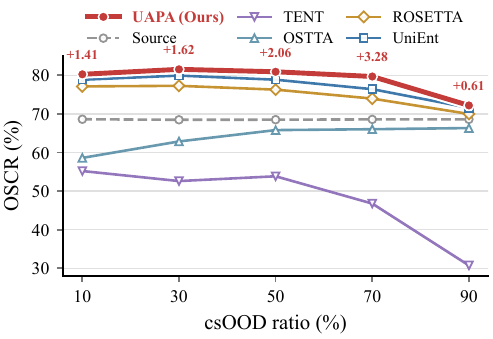}
    \caption{OSCR comparison on CIFAR-10-C under varying csOOD ratios. Values above \methodname{} denote its gains in percentage points over the strongest baseline at each ratio.}
    \label{fig:ood_ratio_sweep}
\end{figure}

\textbf{t-SNE visualization.} To illustrate the effects of different methods on csID classification and csOOD detection, we visualize the feature representations of CIFAR-10-C test samples, with SVHN-C test samples serving as csOOD samples using t-SNE \cite{maaten2008visualizing}, as shown in Fig. \ref{fig:tsne}. As observed, after adaptation with TENT, the feature distributions of known-class (csID) and unknown-class (csOOD) samples are mixed. Although UniEnt shows some improvement in distinguishing csID and csOOD samples, there is still considerable overlap, with some csID samples being mixed with csOOD samples. In contrast, \methodname{} produces more compact csID clusters and a clearer visual separation between csID and csOOD samples. This qualitative observation is consistent with the prototype-relative feature regularization introduced by CPEA, which attracts csID features toward their corresponding target prototypes and reduces the maximum similarity between csOOD features and known-class prototypes.

\textbf{Comparison with GMM-based ID/OOD Separation.}
To evaluate our ID/OOD separation mechanism, we construct a variant that replaces it with a UniEnt-style Gaussian Mixture Model (GMM) gate while keeping all other components unchanged. As shown in Table~\ref{tab:gmm_result}, the GMM-based variant outperforms the Source and TENT baselines across all metrics, showing that the overall framework remains effective with an alternative separation strategy. Nevertheless, the full model achieves better results than the GMM-based variant across all metrics on both CIFAR-10-C and CIFAR-100-C. On average, the proposed separation mechanism reduces FPR@95TPR by 5.32 percentage points and improves OSCR by 0.76 percentage points over the GMM-based alternative. These results suggest that combining the classifier-prototype score with the mini-batch-normalized evidence-derived uncertainty score yields a more effective csID/csOOD partition than the evaluated GMM-based alternative under the current protocol.

\textbf{Robustness to varying csOOD ratios.}
To examine the sensitivity of \methodname{} to open-set stream composition, we vary the csOOD proportion in each test mini-batch from 10\% to 90\% on CIFAR-10-C while keeping the remaining settings unchanged. As shown in Fig.~\ref{fig:ood_ratio_sweep}, \methodname{} achieves the highest OSCR at every evaluated ratio. Its gains over the strongest competing method are 1.41, 1.62, 2.06, 3.28, and 0.61 percentage points at csOOD ratios of 10\%, 30\%, 50\%, 70\%, and 90\%, respectively. The performance of \methodname{} remains stable from 10\% to 70\% csOOD. Although its OSCR decreases when csOOD samples dominate the stream at a ratio of 90\%, it still performs best among the compared methods. These results show that \methodname{} remains effective across varying csID/csOOD compositions, including highly imbalanced test streams.

\begin{table*}[t]
\centering
\caption{Plug-and-play performance of~\methodname{} on CIFAR-10-C, CIFAR-100-C, and their average results.}
\label{tab:plugin_cifar}
\renewcommand{\arraystretch}{1.0}
\setlength{\tabcolsep}{4pt}
\begin{tabular}{lcccccccccccc}
\toprule
\multirow{2}{*}{\textbf{Method}} 
& \multicolumn{4}{c}{\textbf{CIFAR-10-C}} 
& \multicolumn{4}{c}{\textbf{CIFAR-100-C}} 
& \multicolumn{4}{c}{\textbf{Average}} \\
\cmidrule(lr){2-5} \cmidrule(lr){6-9} \cmidrule(lr){10-13}
& Acc$\uparrow$ & AUROC$\uparrow$ & FPR@95TPR$\downarrow$ & OSCR$\uparrow$
& Acc$\uparrow$ & AUROC$\uparrow$ & FPR@95TPR$\downarrow$ & OSCR$\uparrow$
& Acc$\uparrow$ & AUROC$\uparrow$ & FPR@95TPR$\downarrow$ & OSCR$\uparrow$ \\
\midrule
TENT      
& 79.12 & 67.79 & 95.76 & 58.27 
& 50.19 & 59.42 & 93.54 & 36.25 
& 64.66 & 63.61 & 94.65 & 47.26 \\
\rowcolor{pink!30}+ \methodname
& \textbf{84.53} & \textbf{96.18} & \textbf{13.57} & \textbf{82.27} 
& \textbf{60.14} & \textbf{93.26} & \textbf{27.59} & \textbf{58.08} 
& \textbf{72.34} & \textbf{94.72} & \textbf{20.58} & \textbf{70.18} \\
\midrule
EATA      
& 76.17 & 55.43 & 92.88 & 46.66 
& 61.34 & 87.05 & 56.06 & 57.27 
& 68.76 & 71.24 & 74.47 & 51.97 \\
\rowcolor{pink!30}+ \methodname
& \textbf{84.69} & \textbf{92.75} & \textbf{27.24} & \textbf{80.23} 
& \textbf{61.51} & \textbf{92.16} & \textbf{34.30} & \textbf{59.07} 
& \textbf{73.10} & \textbf{92.46} & \textbf{30.77} & \textbf{69.65} \\
\midrule
COME      
& \textbf{85.31} & 75.34 & 84.01 & 69.29 
& 58.09 & 72.28 & 79.24 & 47.49 
& 71.70 & 73.81 & 81.63 & 58.39 \\
\rowcolor{pink!30}+ \methodname
& 84.87 & \textbf{79.53} & \textbf{73.28} & \textbf{71.54} 
& \textbf{59.93} & \textbf{76.56} & \textbf{78.61} & \textbf{51.48} 
& \textbf{72.40} & \textbf{78.05} & \textbf{75.95} & \textbf{61.51} \\
\midrule
OSTTA     
& 84.24 & 72.67 & 76.98 & 64.98 
& 60.03 & 75.09 & 82.78 & 51.23 
& 72.14 & 73.88 & 79.88 & 58.11 \\
\rowcolor{pink!30}+ \methodname
& \textbf{84.55} & \textbf{92.58} & \textbf{26.15} & \textbf{79.81} 
& \textbf{60.59} & \textbf{89.54} & \textbf{38.48} & \textbf{57.15} 
& \textbf{72.57} & \textbf{91.06} & \textbf{32.32} & \textbf{68.48} \\
\bottomrule
\end{tabular}
\end{table*}

\begin{table}[t]
\renewcommand{\arraystretch}{1.0}
\renewcommand{\tabcolsep}{4.0pt}
\begin{center}
\caption{Results of different methods on Tiny-ImageNet-C with ResNet-50 backbone. We report Accuracy (Acc), AUROC, FPR@95TPR, OSCR, and the per-image additional inference time introduced by \methodname{} (in milliseconds).}
\label{tab:plug-and-plug}
\begin{tabular}{lccccc}
\toprule
\textbf{Method}              & {\makecell[c]{Acc$\uparrow$}} & {\makecell[c]{AUROC$\uparrow$}} & {\makecell[c]{FPR@95TPR$\downarrow$}}& {\makecell[c]{OSCR$\uparrow$}}  &Time (ms)$\downarrow$\\ \midrule
TENT~\cite{wang2021tent}              & 25.45 & 52.22 & 94.29    & 17.48 &-\\
\rowcolor{pink!30}+ \methodname       & \textbf{33.85} & \textbf{67.47} & \textbf{81.06} & \textbf{28.10} &+0.22\\ \hline
EATA~\cite{niu2022eata}             & 34.46 & 55.86 & 92.52    & 25.92 &-\\
\rowcolor{pink!30}+ \methodname       & \textbf{35.47} & \textbf{56.94} & \textbf{88.72}    &\textbf{27.43} &+0.26\\ \hline
OSTTA~\cite{lee2023towards}  & 34.75 & 53.00   & 95.29    & 25.54 &-\\
\rowcolor{pink!30}+ \methodname       &\textbf{34.87} & \textbf{53.34} & \textbf{92.95}    & \textbf{25.68} &+0.31\\ \hline
COME~\cite{zhangcome}   & 34.28 & 49.27 & 93.92    & 23.65&- \\
\rowcolor{pink!30}+ \methodname      & \textbf{35.19} & \textbf{55.61} & \textbf{88.80}    & \textbf{26.56} &+0.33\\
\bottomrule
\end{tabular}
\end{center}
\end{table}

\textbf{Plug-and-Play Compatibility of \methodname{}.}
To assess the plug-and-play compatibility of our proposed \methodname{}, we incorporate it into four representative adaptation methods, including TENT~\cite{wang2021tent}, EATA~\cite{niu2022eata}, OSTTA~\cite{lee2023towards}, and COME~\cite{zhangcome}.
For each baseline, we retain its original adaptation objective, update strategy, and method-specific components, while incorporating \methodname's sample separation, prototype refinement, and contrastive regularization, unless otherwise stated.
As shown in Fig.~\ref{fig:existing_with-ours}, Table~\ref{tab:plugin_cifar}, and Table~\ref{tab:plug-and-plug}, incorporating \methodname{} improves open-set detection and OSCR across all evaluated baselines. On the CIFAR benchmarks, combining \methodname{} with TENT increases the average OSCR from 47.26\% to 70.18\% and reduces FPR@95TPR from 94.65\% to 20.58\%. It also improves the average OSCR of EATA, COME, and OSTTA by 17.68, 3.12, and 10.37 percentage points, respectively. Although the magnitude of improvement varies across baselines, the consistent gains indicate that \methodname{} is compatible with different adaptation objectives. On Tiny-ImageNet-C, \methodname{} also improves the OSCR of all four baselines while introducing only about 0.3 ms of additional processing time per image. In particular, when combined with TENT, it improves OSCR by 10.62 percentage points with an additional latency of 0.22 ms per image. These results indicate a favorable performance--latency trade-off and support the use of \methodname{} as a plug-in module for open-set test-time adaptation.

\begin{table}[t]
\renewcommand{\arraystretch}{1.0}
\renewcommand{\tabcolsep}{8.0pt}
\begin{center}
\caption{Results on Tiny-ImageNet-C with ViT-Base under the single-class-per-batch setting.}
\label{tab:single-class}
\begin{tabular}{lcccc}
\toprule
\textbf{Method}              & {\makecell[c]{Acc$\uparrow$}} & {\makecell[c]{AUROC$\uparrow$}} & {\makecell[c]{FPR@95TPR$\downarrow$}}& {\makecell[c]{OSCR$\uparrow$}} \\ \midrule
Source              & 22.09 & 48.45 & 94.85 & 16.29 \\
BN Adapt~\cite{li2017revisiting} & 22.09 & 51.96 & \underline{94.52} & 16.54 \\
CoTTA~\cite{wang2022cotta} & 15.13 & 42.04 & 95.45 & 9.40 \\
\midrule
TENT~\cite{wang2021tent}              & 14.86 & 47.92 & 96.05 & 10.07 \\
EATA~\cite{niu2022eata}        & 22.26 & 51.98 & 94.64 & 16.64 \\ 
SAR~\cite{niutowards}             & 22.24 & 51.76 & 94.58 & 16.50 \\
COME~\cite{zhangcome}   & 22.34 & \underline{52.05} & \textbf{94.34} & 16.61 \\
OSTTA~\cite{lee2023towards} & \underline{23.05} & 50.83 & 95.32 & \underline{17.00} \\
UniEnt~\cite{gao2024unified} & 22.63 & 52.04 & 94.83 & 16.76 \\
ROSETTA~\cite{zhao2026mitigating} & 22.71 & 49.31 & 96.12 & 16.24 \\
\rowcolor{pink!30}\textbf{\methodname{} (Ours)}       & \textbf{23.31} & \textbf{52.68} & 94.54 & \textbf{17.63} \\
\bottomrule
\end{tabular}
\end{center}
\vspace{-6pt}
\end{table}

\textbf{Robustness under the Single-Class-per-Batch Setting.}
We evaluate the proposed \methodname{} under a single-class-per-batch setting, where each test mini-batch contains samples from only one class. This setting represents an extreme form of batch-level class imbalance, in which each update observes only partial class information and lacks within-batch class diversity. In \methodname{}, prototype similarity is computed using the normalized weights of the pretrained classifier head, which provide class references independent of the class composition of the current mini-batch. Meanwhile, DPR accumulates features from samples identified as csID in class-wise buffers and leaves prototypes of unobserved classes unchanged, thereby avoiding unsupported prototype updates under incomplete batch observations. From Table~\ref{tab:single-class},
\methodname{} achieves the highest Acc, AUROC, and OSCR in this setting. Nevertheless, FPR@95TPR remains high at 94.54\%, indicating that reliable csOOD rejection is still challenging under the single-class-per-batch setting. 
Compared with the second-best results, \methodname{} improves Acc by 0.26 percentage points and OSCR by 0.63 percentage points. The results suggest that the combination of batch-independent classifier prototypes and buffered target-prototype refinement helps \methodname{} remain effective when each mini-batch contains only limited class information.

\begin{figure*}[t]
\centering

\subfloat[Effect of $\lambda$]{
    \begin{tikzpicture}
        \begin{axis}[
            width=0.315\linewidth,
            height=4.45cm,
            xlabel={$\lambda$},
            ylabel={OSCR (\%)},
            xmin=0.5, xmax=1.0,
            ymin=10, ymax=31,
            xtick={0.5,0.6,0.7,0.8,0.9,1.0},
            ytick={10,15,20,25,30},
            axis line style={black!70},
            tick style={black!70},
            tick align=outside,
            grid=major,
            grid style={draw=mygrid, line width=0.25pt},
            label style={font=\small},
            tick label style={font=\footnotesize},
            legend to name=tinyhyperlegend,
            legend columns=2,
            legend style={
                draw=none,
                fill=none,
                font=\scriptsize,
                /tikz/every even column/.append style={column sep=8pt}
            }
        ]
            \addplot[color=myblue, line width=1.0pt, mark=*, mark size=1.9pt] coordinates {
                (0.5,14.42)
                (0.6,15.57)
                (0.7,18.89)
                (0.8,24.42)
                (0.9,28.10)
                (1.0,27.87)
            };
            \addlegendentry{ResNet-50}

            \addplot[color=myred, line width=1.0pt, mark=square*, mark size=1.9pt] coordinates {
                (0.5,18.98)
                (0.6,20.21)
                (0.7,21.58)
                (0.8,21.44)
                (0.9,21.44)
                (1.0,21.44)
            };
            \addlegendentry{ViT-Base}

            \addplot[only marks, mark=o, mark size=4.0pt, line width=1.1pt, color=myblue, fill=white, forget plot] coordinates {(0.9,28.10)};
            \addplot[only marks, mark=square, mark size=4.0pt, line width=1.1pt, color=myred, fill=white, forget plot] coordinates {(0.7,21.58)};
        \end{axis}
    \end{tikzpicture}
}
\hfill
\subfloat[Effect of $\tau$]{
    \begin{tikzpicture}
        \begin{axis}[
            width=0.315\linewidth,
            height=4.45cm,
            xlabel={$\tau$},
            xmin=0.54, xmax=1.01,
            ymin=10, ymax=31,
            xtick={0.55,0.65,0.75,0.85,0.95},
            ytick={10,15,20,25,30},
            axis line style={black!70},
            tick style={black!70},
            tick align=outside,
            grid=major,
            grid style={draw=mygrid, line width=0.25pt},
            label style={font=\small},
            tick label style={font=\footnotesize}
        ]
            \addplot[color=myblue, line width=1.0pt, mark=*, mark size=1.9pt] coordinates {
                (0.55,26.09)
                (0.60,27.37)
                (0.65,29.18)
                (0.70,28.10)
                (0.75,27.19)
                (0.80,20.65)
                (0.90,16.27)
                (1.00,16.72)
            };

            \addplot[color=myred, line width=1.0pt, mark=square*, mark size=1.9pt] coordinates {
                (0.55,13.47)
                (0.60,12.61)
                (0.65,12.94)
                (0.70,11.73)
                (0.75,16.03)
                (0.80,15.41)
                (0.85,18.37)
                (0.90,21.58)
                (0.95,21.53)
                (1.00,21.44)
            };

            \addplot[only marks, mark=o, mark size=4.0pt, line width=1.1pt, color=myblue, fill=white, forget plot] coordinates {(0.65,29.18)};
            \addplot[only marks, mark=square, mark size=4.0pt, line width=1.1pt, color=myred, fill=white, forget plot] coordinates {(0.90,21.58)};
        \end{axis}
    \end{tikzpicture}
}
\hfill
\subfloat[Effect of $\gamma$]{
    \begin{tikzpicture}
        \begin{axis}[
            width=0.315\linewidth,
            height=4.45cm,
            xlabel={$\gamma$},
            xmin=-0.02, xmax=0.62,
            ymin=19, ymax=30,
            xtick={0.0,0.1,0.2,0.3,0.4,0.5,0.6},
            ytick={20,22,24,26,28,30},
            axis line style={black!70},
            tick style={black!70},
            tick align=outside,
            grid=major,
            grid style={draw=mygrid, line width=0.25pt},
            label style={font=\small},
            tick label style={font=\footnotesize}
        ]
            \addplot[color=myblue, line width=1.0pt, mark=*, mark size=1.9pt] coordinates {
                (0.0,27.58)
                (0.1,28.01)
                (0.2,28.10)
                (0.3,27.11)
                (0.4,26.66)
                (0.5,26.20)
                (0.6,26.16)
            };

            \addplot[color=myred, line width=1.0pt, mark=square*, mark size=1.9pt] coordinates {
                (0.0,21.56)
                (0.1,21.58)
                (0.2,21.47)
                (0.3,21.40)
                (0.4,21.24)
                (0.5,21.01)
                (0.6,20.67)
            };

            \addplot[only marks, mark=o, mark size=4.0pt, line width=1.1pt, color=myblue, fill=white, forget plot] coordinates {(0.2,28.10)};
            \addplot[only marks, mark=square, mark size=4.0pt, line width=1.1pt, color=myred, fill=white, forget plot] coordinates {(0.1,21.58)};
        \end{axis}
    \end{tikzpicture}
}

\vspace{1pt}
{\centering
\pgfplotslegendfromname{tinyhyperlegend}\par
}

\caption{Effect of key hyperparameters on OSCR on Tiny-ImageNet-C with ResNet-50 and ViT-Base backbones. We vary one hyperparameter at a time while keeping the others fixed at their selected values.}
\label{fig:hyper_sensitivity_tiny}
\end{figure*}
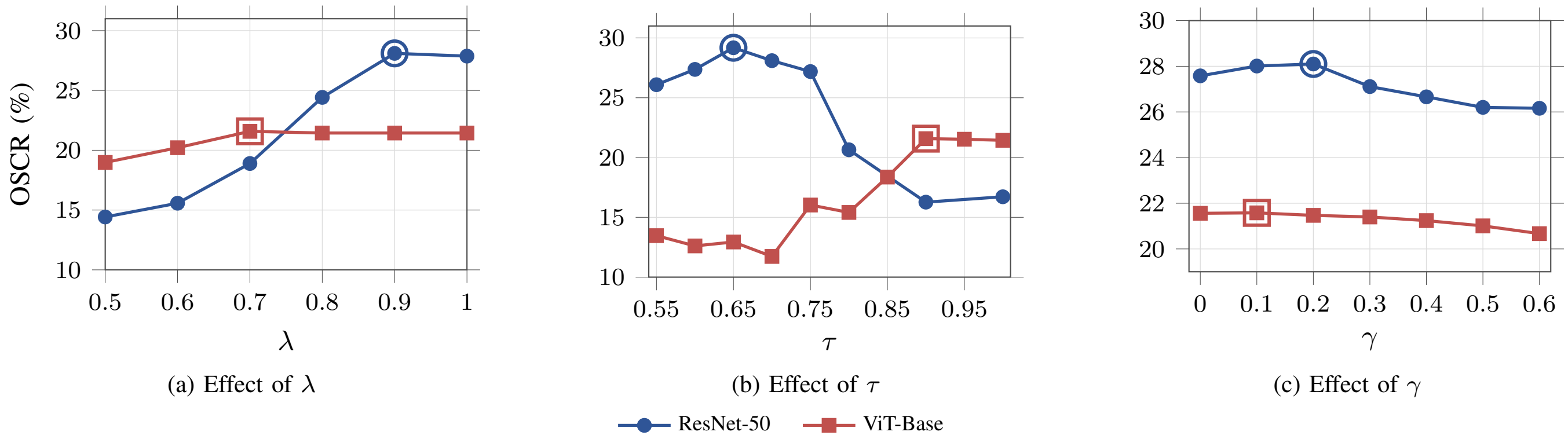

\textbf{Hyperparameter Analysis.}
We examine the effects of the score-balance weight $\lambda$, separation threshold $\tau$, and contrastive-loss weight $\gamma$ on Tiny-ImageNet-C. As shown in Fig.~\ref{fig:hyper_sensitivity_tiny}, \methodname{} maintains competitive performance over a reasonably broad range around the selected values. The contrastive-loss weight $\gamma$ has a relatively mild effect on both backbones. The score-balance weight $\lambda$ exhibits a broad high-performing range, although ResNet-50 benefits more from larger values. The separation threshold $\tau$ shows backbone-dependent preferred ranges because it directly controls the csID/csOOD partition and the resulting score distributions differ across architectures. Overall, the selected configurations lie within stable operating regions rather than isolated optima.

\section{Conclusion}

In this paper, we propose \textbf{U}ncertainty-\textbf{A}ware \textbf{P}rototype Learning for Open-Set Test-Time \textbf{A}daptation (\textbf{\methodname}), a plug-and-play method for test streams containing both csID and csOOD samples. Specifically, \methodname{} combines a classifier-prototype score with a mini-batch-normalized evidence-derived uncertainty score to separate csID and csOOD samples. \methodname{} further employs Dynamic Prototype Refinement to update class-specific target prototypes from buffered features of selected csID samples. Contrastive Prototype Entropy Adaptation then combines class-conditional entropy optimization with prototype-level contrastive learning to increase pseudo-label-relative similarity margins for csID features and reduce the maximum similarity between csOOD features and initialized known-class prototypes. Experimental results on CIFAR-C and Tiny-ImageNet-C with both CNN and ViT backbones demonstrate the effectiveness of \methodname{}, consistently improving open-set classification and detection with limited additional computational cost.

\small
\bibliographystyle{IEEEtran}
\bibliography{reference}

\end{document}